%% file: sample-sigconf.tex
\documentclass[sigconf]{acmart}

\input{template-variables}
\input{packages}
\input{macros}

\AtBeginDocument{%
  \providecommand\BibTeX{{%
    Bib\TeX}}}

\copyrightyear{2022}
\acmYear{2022}
\setcopyright{rightsretained}

\acmConference[KDD '22]{Proceedings of the 28th ACM SIGKDD Conference on Knowledge Discovery and Data Mining}{August 14--18, 2022}{Washington, DC, USA}
\acmBooktitle{Proceedings of the 28th ACM SIGKDD Conference on Knowledge Discovery and Data Mining (KDD '22), August 14--18, 2022, Washington, DC, USA}
\acmISBN{978-1-4503-9385-0/22/08}
\acmDOI{10.1145/3534678.3539203}

\usepackage{etoolbox}
\makeatletter
\patchcmd{\maketitle}{\@copyrightpermission}{
   \begin{minipage}{0.3\columnwidth}
     \href{https://creativecommons.org/licenses/by/4.0/}{\includegraphics[width=0.90\textwidth]{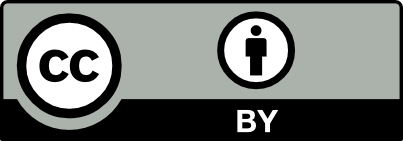}}
   \end{minipage}\hfill
   \begin{minipage}{0.7\columnwidth}
     \href{https://creativecommons.org/licenses/by/4.0/}{This work is licensed under a Creative Commons Attribution International 4.0 License.}
   \end{minipage}

   \vspace{5pt}
}{}{}

\makeatother

\acmSubmissionID{2604}

\begin{document}

\title{GraphWorld: Fake Graphs Bring Real Insights for GNNs}

\input{authors}

\renewcommand{\shortauthors}{John Palowitch et al.}
%% No italics and no comma

\input{contents/0.abstract}



\input{contents/0.teaserfigure.3d}

\maketitle
\input{contents/1.introduction}
\input{contents/2.related-work}
\input{contents/3.solution}
\input{contents/4.experiments}
\input{contents/5.conclusion}
\balance
\bibliographystyle{ACM-Reference-Format}
\bibliography{acmart}

\newpage
\appendix
\input{contents/999.appendix}

\end{document}

%% file: template-variables.tex
\AtBeginDocument{%
  \providecommand\BibTeX{{%
    \normalfont B\kern-0.5em{\scshape i\kern-0.25em b}\kern-0.8em\TeX}}}

\acmDOI{}
\copyrightyear{2021}
\acmYear{2021}
\setcopyright{iw3c2w3}
\acmConference[KDD '22]{Proceedings of ACM SIGKDD Conference on Knowledge Discovery and Data Mining 2022}{August 14--18, 2022}{Washington DC, USA}
\acmBooktitle{Proceedings of ACM SIGKDD Conference on Knowledge Discovery and Data Mining 2022 (KDD '22), August 14--18, Washington DC, USA}
\acmPrice{}
\acmISBN{}

%% file: packages.tex
\usepackage[utf8]{inputenc} % allow utf-8 input
\usepackage[T1]{fontenc}    % use 8-bit T1 fonts
\usepackage{amsfonts}       % blackboard math symbols
\usepackage{amsmath}        % math environments
\usepackage{thmtools}
\usepackage{nicefrac}       % compact symbols for 1/2, etc.
\usepackage{pifont}         % for ticks
\usepackage{xspace}         % convenience space

\usepackage{hyperref}       % hyperlinks
\usepackage{url}            % simple URL typesetting

\usepackage{booktabs}       % professional-quality tables
\usepackage{tabularx}
\usepackage{multirow}
\usepackage{balance}
\usepackage{enumitem}

\usepackage{hhline}

\usepackage{algorithm}      % algorithm
\usepackage[noend]{algpseudocode} % algorithm

\usepackage[utf8]{inputenc} 
\usepackage[T1]{fontenc}

\usepackage{siunitx}
\usepackage{graphicx}       % images
\usepackage{subfig}
\usepackage{tikz}
\usepackage{pgfplots}
\pgfplotsset{compat=1.16}
\usepgfplotslibrary{fillbetween, groupplots, colorbrewer}

\usepackage{xspace}

%% file: macros.tex
\definecolor{cycle1}{RGB}{155, 180, 14}
\definecolor{cycle2}{RGB}{0, 84, 186}
\definecolor{cycle3}{RGB}{147, 0, 0}
\definecolor{cycle4}{RGB}{0, 187, 104}
\definecolor{cycle5}{RGB}{225, 133, 240}
\definecolor{cycle6}{RGB}{134, 80, 0}
\definecolor{cycle7}{RGB}{0, 192, 211}
\definecolor{cycle8}{RGB}{255, 118, 150}
\definecolor{cycle9}{RGB}{2, 89, 0}
\definecolor{cycle10}{RGB}{0, 150, 227}
\definecolor{cycle11}{RGB}{234, 119, 47}
\definecolor{cycle12}{RGB}{168, 0, 102}
\definecolor{cycle13}{RGB}{0, 139, 121}
\definecolor{cycle14}{RGB}{171, 133, 0}
\definecolor{cyclegray}{RGB}{189,189,189}

\DeclareMathOperator*{\argmax}{\arg\max}

\newcommand*{\gw}{\texttt{GraphWorld}\xspace}

%% file: authors.tex
% \author{John Palowitch, Anton Tsitsulin, Brandon Mayer, Bryan Perozzi}
% \email{ { palowitch,tsitsulin,bmayer}@google.com, bperozzi@acm.org}

\author{John Palowitch}
\email{palowitch@google.com}
\affiliation{%
  \institution{Google Research}
  \country{USA}
}
\author{Anton Tsitsulin}
\email{tsitsulin@google.com}
\affiliation{%
  \institution{Google Research}
  \country{USA}
}
\author{Brandon Mayer}
\email{bmayer@google.com}
\affiliation{%
  \institution{Google Research}
  \country{USA}
}
\author{Bryan Perozzi}
\email{bperozzi@acm.org}
\affiliation{%
  \institution{Google Research}
  \country{USA}
}

%% file: contents/0.abstract.tex
\begin{abstract}
Despite advances in the field of Graph Neural Networks (GNNs), only a small number ($\texttt{\char`\~}5$) of datasets are currently used to evaluate new models. This continued reliance on a handful of datasets provides minimal insight into the performance differences between models, and is especially challenging for industrial practitioners who are likely to have datasets which are very different from academic benchmarks.
In the course of our work on GNN infrastructure and open-source software at Google, we have sought to develop benchmarks that are robust, tunable, scalable, and generalizable.

In this work we introduce \gw, a novel methodology and system for benchmarking GNN models on an arbitrarily-large population of \emph{synthetic} graphs for \emph{any} conceivable GNN task.
\gw allows a user to efficiently generate a \emph{world} with millions of statistically diverse datasets. It is accessible, scalable, and easy to use. \gw can be run on a single machine without specialized hardware, or it can be easily scaled up to run on arbitrary clusters or cloud frameworks. Using \gw, a user has fine-grained control over graph generator parameters, and can benchmark arbitrary GNN models with built-in hyperparameter tuning.

We present insights from \gw experiments on the performance characteristics of thirteen GNN models and baselines over millions of benchmark datasets. We show that \gw efficiently explores regions of benchmark dataset space uncovered by standard benchmarks, revealing comparisons between models that have not been historically obtainable. Using \gw, we also are able to study in-detail the relationship between graph properties and task performance metrics, which is nearly impossible with the classic collection of real-world benchmarks.
\end{abstract}

%% file: contents/0.teaserfigure.3d.tex
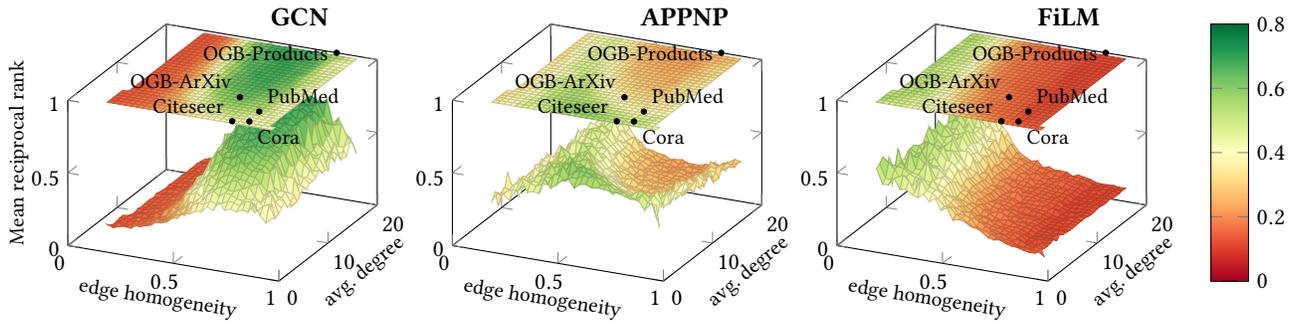
\begin{teaserfigure}
\centering
\begin{tikzpicture}
\begin{groupplot}[group style={
    group name=myplot,
    group size= 3 by 1,
    horizontal sep=1cm,
    vertical sep=1cm},
    height=5cm,
  width=.32\linewidth,
  colormap/RdYlGn,
  mesh/ordering=y varies,
  mesh/rows=30,
  unbounded coords=jump,
  zmin=0,
  zmax=1,
  xmin=0,
  xmax=1,
  point meta min=0,
  point meta max=.8,
  ymin=0,
  label style={font=\small},
  xlabel=edge homogeneity,
  title style={font=\bfseries\large},
  ylabel=avg.\ degree,
  xlabel style={at={(axis description cs:0.3,0)},rotate=-9,anchor=north},
  ylabel style={at={(axis description cs:.92,.09)},rotate=40,anchor=north},
  zlabel style={at={(axis description cs:-0.1,0.5)},anchor=south},
      title style={at={(0.75,.9,0.6)}},
    ]
\nextgroupplot[
      title=GCN,
  zlabel=Mean reciprocal rank,]
  \addplot3[surf] table {data/kdd_bucketed/mode2_sbm_GCN.csv};
  \addplot3[surf, point meta=explicit] table [z expr=1, meta index=2] {data/kdd_bucketed/mode2_sbm_GCN.csv};
  \addplot3 [color=black, draw=none, mark=*, mark size=1]
            table[row sep=crcr] {%
            0.8138 2 1\\ % cora
            0.7461 1.42 1\\ % citeseer
            0.6551 6.89 1\\ % arxiv
            0.8024 4.50 1\\ % pubmed
            0.8076 20 1\\ % products
            };
    \node[anchor=north west] at (axis cs: 0.8138,2,1) {\small Cora};
    \node[anchor=south east] at (axis cs: 0.7461,1.42,1) {\small Citeseer};
    \node[anchor=south east] at (axis cs: 0.655,6.89,1) {\small OGB-ArXiv};
    \node[anchor=south west] at (axis cs: 0.8024,4.50,1) {\small PubMed};
    \node[anchor=east] at (axis cs: 0.8076,20,1) {\small OGB-Products};
\nextgroupplot[
      title=APPNP,]
  \addplot3[surf] table {data/kdd_bucketed/mode2_sbm_APPNP.csv};
  \addplot3[surf, point meta=explicit] table [z expr=1, meta index=2] {data/kdd_bucketed/mode2_sbm_APPNP.csv};
  \addplot3 [color=black, draw=none, mark=*, mark size=1]
            table[row sep=crcr] {%
            0.8138 2 1\\ % cora
            0.7461 1.42 1\\ % citeseer
            0.6551 6.89 1\\ % arxiv
            0.8024 4.50 1\\ % pubmed
            0.8076 20 1\\ % products
            };
    \node[anchor=north west] at (axis cs: 0.8138,2,1) {\small Cora};
    \node[anchor=south east] at (axis cs: 0.7461,1.42,1) {\small Citeseer};
    \node[anchor=south east] at (axis cs: 0.655,6.89,1) {\small OGB-ArXiv};
    \node[anchor=south west] at (axis cs: 0.8024,4.50,1) {\small PubMed};
    \node[anchor=east] at (axis cs: 0.8076,20,1) {\small OGB-Products};
\nextgroupplot[
      title=FiLM,
colorbar,
]
  \addplot3[surf] table {data/kdd_bucketed/mode2_sbm_FiLM.csv};
  \addplot3[surf, point meta=explicit] table [z expr=1, meta index=2] {data/kdd_bucketed/mode2_sbm_FiLM.csv};
  \addplot3 [color=black, draw=none, mark=*, mark size=1]
            table[row sep=crcr] {%
            0.8138 2 1\\ % cora
            0.7461 1.42 1\\ % citeseer
            0.6551 6.89 1\\ % arxiv
            0.8024 4.50 1\\ % pubmed
            0.8076 20 1\\ % products
            };
    \node[anchor=north west] at (axis cs: 0.8138,2,1) {\small Cora};
    \node[anchor=south east] at (axis cs: 0.7461,1.42,1) {\small Citeseer};
    \node[anchor=south east] at (axis cs: 0.655,6.89,1) {\small OGB-ArXiv};
    \node[anchor=south west] at (axis cs: 0.8024,4.50,1) {\small PubMed};
    \node[anchor=east] at (axis cs: 0.8076,20,1) {\small OGB-Products};
\end{groupplot}
\end{tikzpicture}
\vspace*{-2ex}
\caption{
\gw uses synthetic data to expose fundamental differences between GNN convolutions.
Shown here are the relative performance results of GCN \cite{kipf2016semi}, APPNP \cite{klicpera2018predict}, and FiLM \cite{brockschmidt2020gnn} across 50,000 synthetic node classification tasks. 
The $x$ and $y$ axes group the synthetic graphs by their structural properties, while the $z$-axis shows the mean reciprocal-rank (MRR) relative to other baselines (Section \ref{ss:models}). 
Standard GNN benchmark datasets (Cora, OGB, etc.) exist in graph property space where model rankings do not change.
\gw discovers graphs which reveal new insights about GNN architectures.
\label{fig:surfaceplots}}
\end{teaserfigure}

%% file: contents/1.introduction.tex
\section{Introduction}
Graph Neural Networks (GNNs) have extended the benefits of deep learning to the non-Euclidean domain, allowing for standardized and re-usable machine learning approaches to problems that involve relational (graph-structured) data \citep{wu2020comprehensive}. GNNs now admit an extremely wide range of architectures and possible tasks, including node classification, whole-graph classification, and link prediction \citep{chami2020machine}. With this growth has come increased calls for proper GNN experimental design \citep{shchur2018pitfalls,you2020design}, refreshed benchmark datasets \citep{hu2020open}, and fair comparisons of GNN models in reproducible settings \citep{errica2019fair,marareproducible}. 

Despite the proliferation of new GNN models, only a few hand-picked benchmarked datasets are currently used to evaluate them \cite{hu2020open}.
The limited scope of these datasets introduces a number of problems.
First, it makes it hard for practitioners to infer which models will generalize well to unseen datasets.
Second, new architectures are proposed only when they beat existing methods on these datasets, which can cause \emph{architectural overfitting} \cite{recht2018cifar,nie2019adversarial}.
Finally, recent efforts at expanding the diversity of available GNN benchmark data have focused on size -- increasing the cost of evaluating models without significantly increasing the variation of graphs considered.

In this work, we introduce \gw, the first tunable, scalable, and reproducible method for analyzing the performance of GNN models \emph{on synthetic benchmark data} for \emph{any given GNN task} (i.e.\ all \{un/semi\}supervised node/graph problems \cite{wu2020comprehensive}). With its ability to generate a vast and diverse ``world" of graph datasets for any specified task, \gw allows comparisons between GNN models and architectures that are not possible with the handful of standard graph datasets on which the current literature depends. As seen in Figure \ref{fig:surfaceplots}, GNN models change sharply in performance ranking when tested on \gw synthetic datasets that are \emph{distant} in graph property space from standard real-world datasets.

\gw is directly motivated by challenges observed in developing and applying GNNs to a wide variety of datasets at Google.
Synthetic datasets have an additional advantage for industrial use -- as there are no underlying privacy concerns, the data is easy to share.
This can help facilitate new model development.
We currently use \gw for model understanding, and will soon be incorporating it into the newly-released TF-GNN package \cite{tfgnn}, as well as in internal systems for testing GNN modules.

This paper's contributions are:

\begin{enumerate}[leftmargin=*]
    \item \textbf{Problem Formulation}. We pose the question: how can we measure GNN model performance across graph datasets with high statistical variance? Modern benchmark datasets, however well-maintained, can be limited in scope, and can be computationally inaccessible to the average researcher.
    
    \item \textbf{Methodology}.  We provide \gw, a graph sampling and GNN training procedure which is capable of testing state-of-the-art GNNs on task datasets beyond the scope of any existing benchmarks.
    
    \item \textbf{Insights}.  We use \gw to conduct large-scale experimental study on over 1 million graph datasets for each of three GNN tasks -- node classification, link prediction, and graph property prediction. We provide a novel method to explore the GNN model performance across all locations in the graph worlds that we generate.
\end{enumerate}

Importantly, the \gw methodology enables experimental analysis of GNNs across \emph{any} task that can be expressed in the theoretical framework we introduce in Section \ref{s:gw}, including tasks not mentioned above such as node regression. Therefore, particular random graph models for synthetic dataset generation are not the focus of our investigation.

The rest of this paper is as follows.
First we discuss recent work and current challenges in benchmarking GNNs. 
We then formally propose \gw as a novel benchmarking system that features many advantages unavailable to GNN experiments that depend only on natural datasets from the literature. 
Next, in Section 4, we demonstrate \gw on node classification, link prediction, and graph property prediction tasks, and show brand-new aggregate performance metrics on GNN models, some of which raise surprising conclusions about prior work and common-sense intuitions.
We close with a discussion about the \gw platform, including its scalability and accessibility compared with large-scale experiments on real-world graphs.

%% file: contents/2.related-work.tex
\section{GNN Experiments: Past and Future}\label{s:motivation}
Progress on GNN architectures is recorded, in large part, by comparing the empirical performance of proposed and existing architectures on particular tasks. In these empirical studies, each task is associated with a number of graph datasets on which GNN models attempt to perform well for the given task. For example, the CORA dataset is a graph of research papers connected by citation edges, curated from online records, on which GNNs must classify the papers (nodes) into research fields (categorical labels). In this case, if a novel GNN model statistically outperforms previously proposed models on the node classification task, this would usually be seen as an indicator that the model makes progress in the field.

\begin{figure}[!tbp]
\includegraphics[width=0.8\columnwidth]{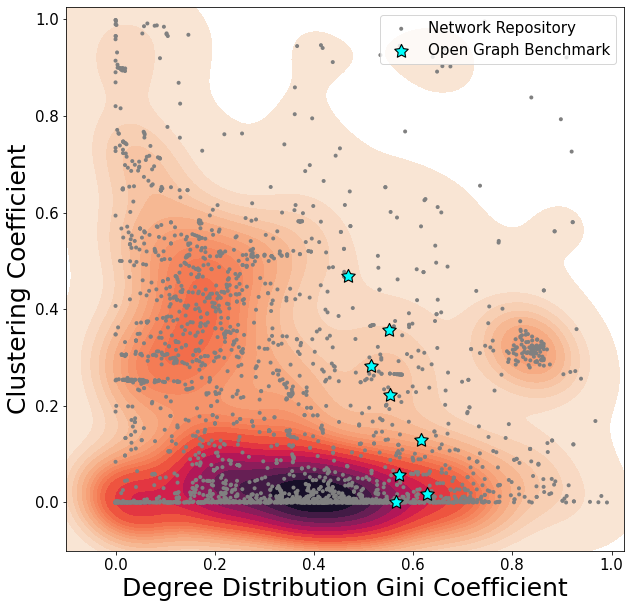}
\vspace*{-4mm}
\caption{\label{fig:cc-vs-gini} The distribution of graphs from the Open Graph Benchmark does not match the general population. Our proposed method, \gw, allows statistically sound insights about GNN model performance via extensive, scalable sampling from an expansive distribution of graphs.}
\vspace*{-5mm}
\end{figure}

In general, such experiments are used to arrive at some insights about how new architectures will perform in realistic scenarios. Most studies feature 1-3 tasks and 1-5 datasets per-task. Note that some datasets contain many graphs, such as a dataset containing molecules represented as atom graphs \citep{morris2020tudataset}.
Multiple-graph datasets have primarily been used to train GNNs on the task of classifying whole graphs, or estimating some property of them. Such datasets can be considered equivalent to a dataset with only one (usually much larger) graph, given that GNN performance on each dataset is measured by a single task and a single set of metrics. 
Even when the dataset collection itself is large, such as TUDataset \cite{morris2020tudataset}, which currently contains more than 130 graph collections, researchers can not report the results on all these datasets due to the lack of space, limiting the reporting to 3--6 most common ones.

These concerns are not specific to the field of GNNs --- the broader machine learning community has identified problems in benchmarking protocols and reporting in other subfields \citep{lipton2018troubling,dacrema2019we,recht2018cifar}.
New benchmarks are being actively created in nearly all areas of machine learning \cite{koh2021wilds, wolfson2020break,cobbe2020leveraging,shridhar2020alfred}.
In the field of GNNs, recent comparative benchmarking studies \cite{khosla2019comparative,luzhnica2019graph,errica2019fair,zhao2020pipeline,dwivedi2020benchmarking} limit themselves to just a few datasets, mainly targeting making fair experimental settings and performing fair hyperparameter tuning.

Possibly due to the existence of well-studied random graph models such as the Stochastic Block Model  \citep{karrer2011stochastic,abbe2017community}, there has been a very recent trend of featuring small synthetic datasets in GNN research, to tease apart model differences that would be harder to observe on standard datasets \cite{tsitsulin2020graph,dwivedi2020benchmarking,tsitsulin2021synthetic,shah2020scale,zhu2020beyond,chen2020can}. However, to date, there is no generalized methodology for producing synthetic, tunable \emph{populations} of GNN task datasets at-scale, nor a concept of how to analyze GNN performance on such populations. This is the main problem we aim to solve with \gw.

\begin{figure}[t]
%\vspace{-0.7cm}
\centering
\subfloat[Unbalanced clusters, $\nicefrac{p}{\!q}=25.0$]{\includegraphics[width=0.2\textwidth]{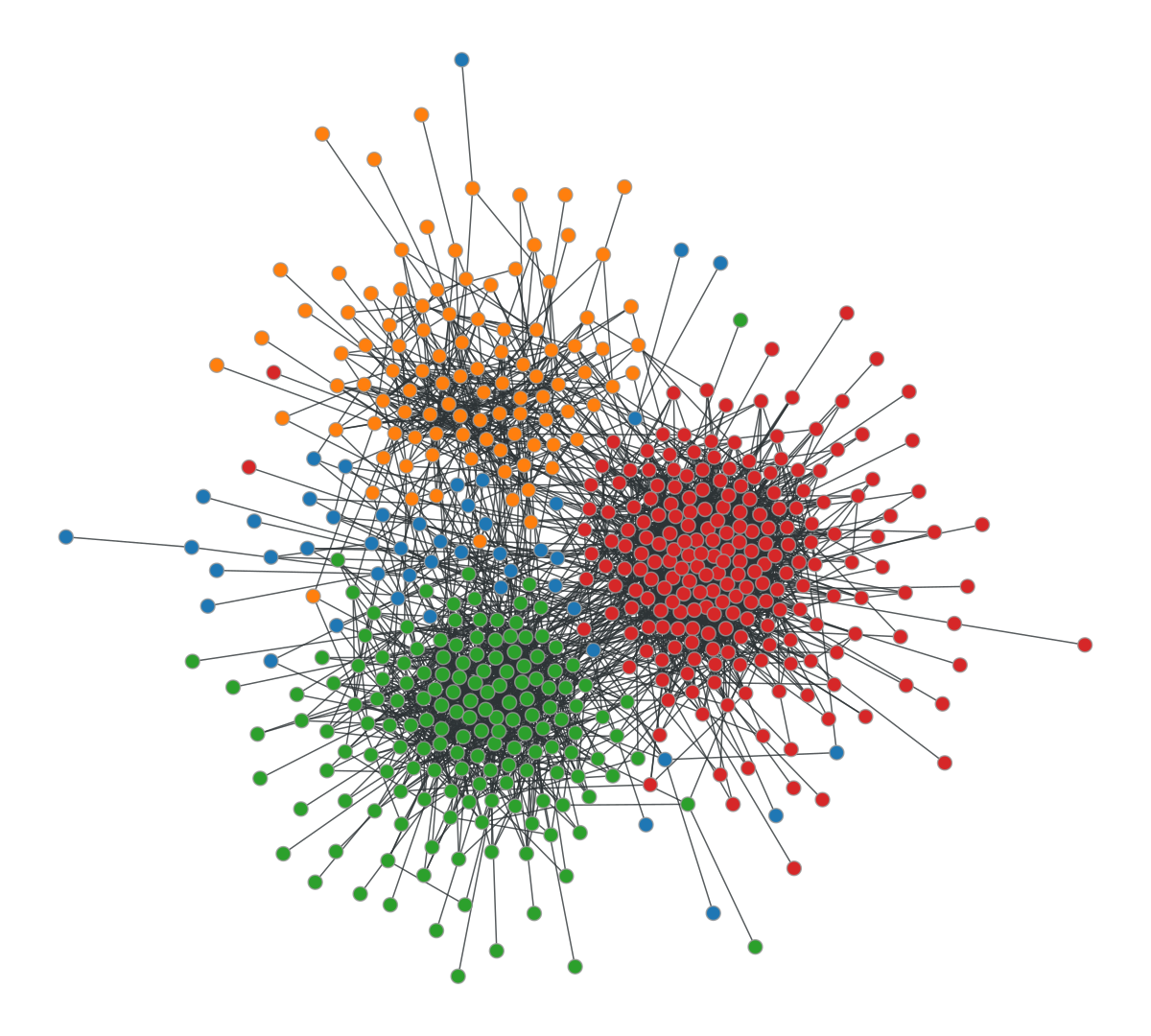}}
\hspace{2mm}
\subfloat[Features PCA, center distance $=0.05$]{\includegraphics[width=0.21\textwidth,trim={0 0 0 0.9cm },clip]{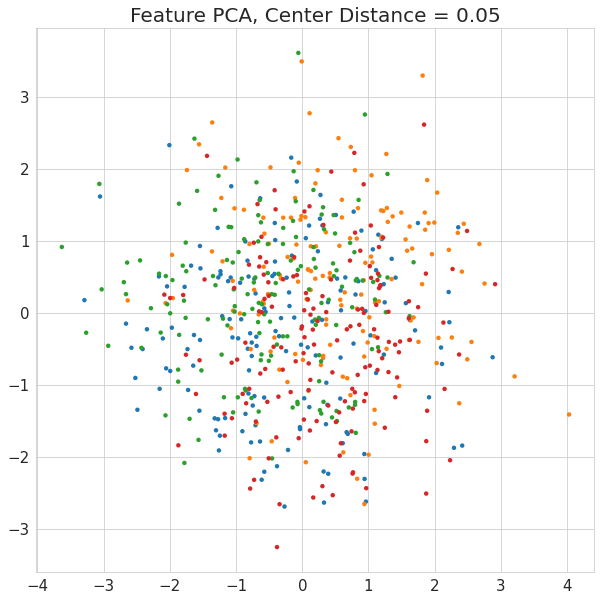}} 
\vspace{-2mm}
\subfloat[Balanced clusters, $\nicefrac{p}{\!q}=5.0$]{\includegraphics[width=0.2\textwidth]{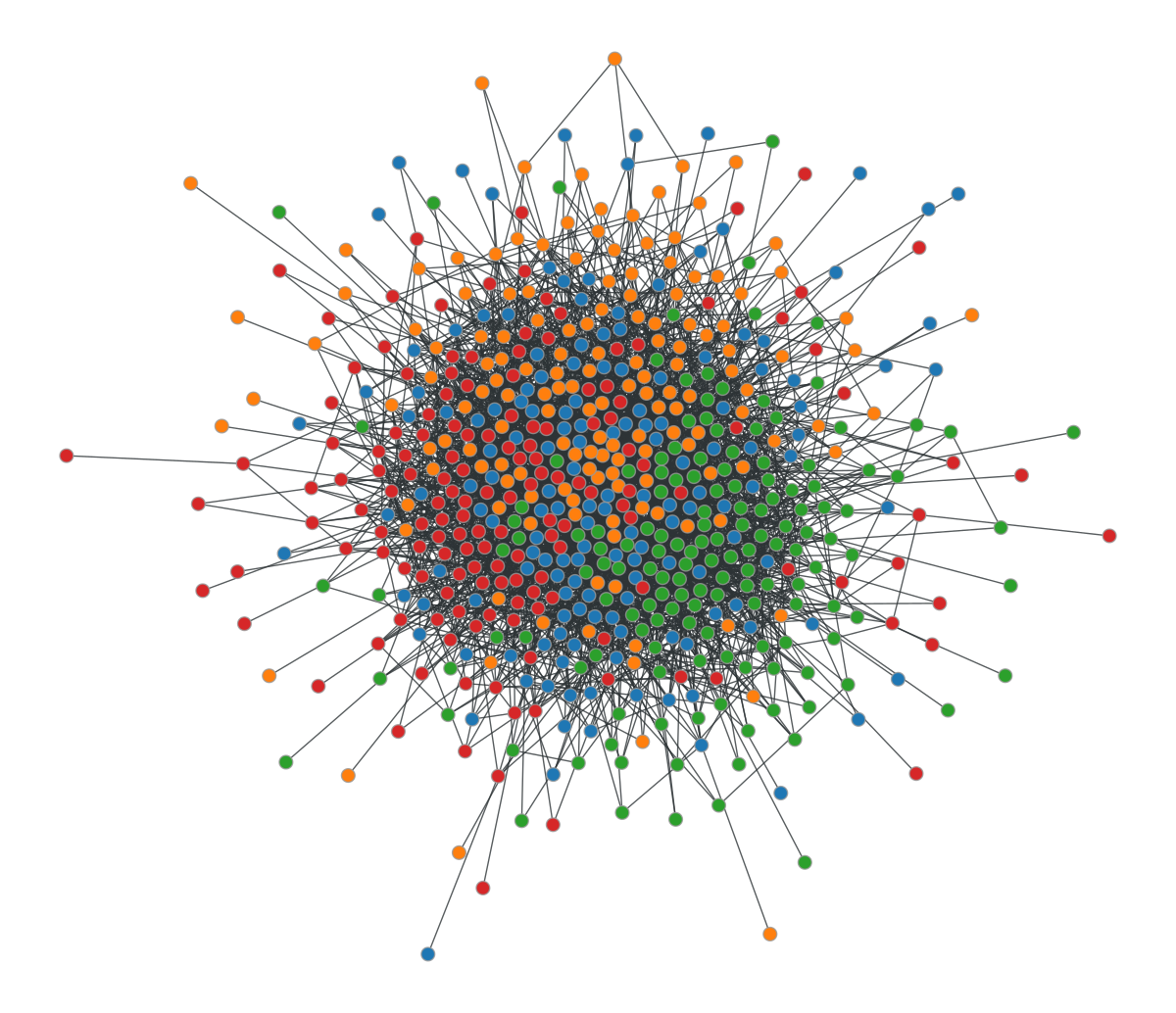}}
\hspace{2mm}
\subfloat[Features PCA, center distance $=3.0$]{\includegraphics[width=0.21\textwidth, trim={0 0 0 0.9cm },clip]{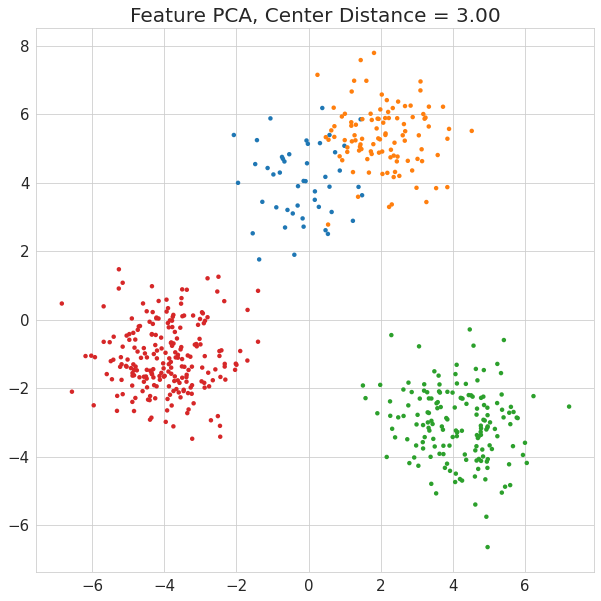}}
\vspace*{-3mm}
\caption{\gw node classification datasets simulated using a Stochastic Block Model with (independent) Gaussian node features. These plots show the effect of toggling \gw generator parameters like graph signal-to-noise ratio ($\nicefrac{p}{\!q}$) and the distance between feature cluster centers.}\label{fig:SBM}
\vspace{-5mm}
\end{figure}

\subsection{Pitfalls of Standard GNN Benchmarks}

In this paper, we question whether or not the status quo of GNN evaluation as described above, is enough to measure progress in the field. While there has been much recent work on improving and standardizing GNN benchmark datasets, relying only on a handful of graph datasets over time is detrimental to the field, for the following three reasons:

\textbf{Inadequate generalization.} Each curated graph dataset represents just one point in the space of all possible datasets that can be associated with the particular GNN task at-hand. The graph (or graphs) in a particular dataset may have properties that favor some GNN models over others, whereas yet-unseen graphs will have different characteristics that could reverse any insights made from the singular trial. Figure \ref{fig:cc-vs-gini} illustrates this point by plotting the joint distribution of the average node clustering coefficient and the degree distribution gini coefficient, for each graph dataset from two populations. The grey points, and the 2D kernel density estimate contours, represent datasets from the massive Network Repository, an open, online graph dataset collection \citep{networkrepository}. The blue points represent node classification and link prediction datasets from the recent Open Graph Benchmark (OGB) \cite{hu2020open}, a library and dataset collection specifically created for reproducible GNN experimentation. 

Here it is important to note that, unlike graph datasets in the OGB, most Network Repository (NR) graph datasets do not have graph or node labels. This is not necessarily by design. The NR is an open, community-sourced collection of graph datasets, and most contributors either do not provide labels, or they determine that labels are not readily available from the graph data source. This means that most NR datasets are not readily available for GNN experimentation. However, it is worth asking the question: if all NR datasets \emph{did} have labels, would GNN model performance generalize to them? There is not currently a way to answer this.

\textbf{Incremental overfitting.} As it is in many machine learning subfields, GNN task datasets are successively re-used across papers, to accurately measure incremental improvements of new architectures. However, this can easily cause, over time, overfitting of new architectures to the datasets, as observed for NLP tasks \citep{nie2019adversarial} and computer vision tasks \cite{recht2018cifar}. This effect will be especially pronounced if the main collection of benchmark graphs have similar structural and statistical properties. 

\textbf{Un-scalable development.} In recent years, there has been a particular focus on scalability in GNN research. The Open Graph Benchmark \cite{hu2020open} increased the size of experiment-friendly benchmark citation graphs by over 1,000x the number of nodes. From one perspective, this can only be natural as computing capabilities grow and graph-based learning problems become increasingly data-flush.  On the other hand, while the availability of giant graphs is important for testing GNN software, platforms, and model complexity, it is not obvious that giant graphs are needed to test GNN \emph{accuracy} or \emph{scientific usefulness}. As the field's benchmark graphs become ever-larger, standardized graph datasets for testing GNN expressiveness become less accessible to the average researcher. Additionally, with large benchmark datasets, it is nearly impossible to investigate GNN hyperparameter tuning techniques or training variance without access to institution-scale computing resources.

%% file: contents/3.solution.tex
\vspace{-1mm}
\section{GRAPHWORLD}\label{s:gw}

The three problems described in the previous section -- inadequate generalization, incremental overfitting, and un-scalable development -- are \emph{inherent} to the need of GNN researchers to benchmark on datasets sourced from the real world. Any handful of such datasets will be necessarily limited, and it is in the best interest of researchers to re-use well maintained datasets from prior work to show how novel methods improves over existing architectures in the field.  Additionally, studies on very large graphs are necessary to demonstrate the scalability of proposed GNN architectures. Such large graphs are expensive to collect and process, and require specialized hardware or computing clusters to train the models, limiting access to research for under-represented groups.

Given that these problems can not be solved with the status quo approach to benchmark design, a \emph{complementary} system is necessary to fill in the gaps that re-usable benchmark datasets cannot. For this, we propose \gw: a distributed framework for simulating diverse \emph{populations} of GNN benchmark datasets, tuning and testing an arbitrary number of GNN models on the population, and extracting population-level insights from the logs of the system. As we explain in this section, and demonstrate with our experiment results, \gw provides \emph{generalizable} GNN insights in a \emph{scalable} manner that is accessible to researchers with low computational resources. In a discussion further in the paper (Section 6), we elaborate on how \gw and its experimental results help resolve the three aforementioned problems. In the rest of this section we describe the core components of \gw, leaving some specifics and implementation details to the Appendix.

\subsection{Graph Generation}
The core component of \gw is the simulation of GNN test datasets using attributed-graph and label generators. Each test dataset is a realization of a parameterized probability distribution $\mathcal{P}(\pi_1, \pi_2, \ldots)$ on $\mathcal{D} = \mathcal{G}\times\mathcal{F}\times\mathcal{L}$, where $\mathcal{G}$ is a collection of graphs, $\mathcal{F}$ is a collection of features, and $\mathcal{L}$ is a collection of labels. (This formulation supports graph classification datasets, which can be represented as a single graph of disjoint graph examples.) The space of possible test datasets $\mathcal{D}$ is fully determined by regions of generator parameters $\pi_1\in\Pi_1, \pi_2\in\Pi_2, \ldots$ provided by the user to the workers via the manager. Each worker in a \gw pipeline generates a \emph{single} realization $D\in\mathcal{D}$ by sampling a generator parameter set $(\pi_1, \pi_2, \ldots)$, and then sampling a single dataset from $\mathcal{P}(\pi_1, \pi_2, \ldots)$. Note: each dataset sampled is actually a combination $D = [D_{train}, D_{test}]$ of training data and testing data for the GNN models.

\textbf{Example.} We illustrate the simulation component of \gw with a node classification task generation. In this paper, we generate worlds of node classification tasks from samples $G\times L\in\mathcal{G}\times\mathcal{L}$ as a Degree-Corrected Stochastic Block Model (DC-SBM) graphs \cite{abbe2017community}, where cluster assignments double as class labels, and node features $F\in\mathcal{F}$ are drawn from within-cluster multivariate Normals. Each of these distributions have many dependent parameters, such as the $p$-to-$q$ ratio (where $p$ and $q$ are the probabilities of in-cluster, out-cluster edges), the degree power-law $\alpha$, the variance of multivariate Normal centers, the cluster size distribution, and many more that we detail fully in the Appendix. Figure \ref{fig:SBM} contains visualizations of two separate attributed DC-SBM realizations from a graph world. In a full run of \gw, each of these parameters (and more) will be randomly sampled from a wide range, producing a diverse classification of node classification datasets.

\subsection{Training, Testing, and Evaluation}
The \gw method simulates a pre-specified number of GNN test datasets $D_1, D_2, \ldots \in \mathcal{D}$, in the manner described above, and trains and tests an arbitrary list of GNN models on each dataset. Note that the particular task is fully-generalizable beyond node classification; we demonstrate link prediction and graph property prediction in Section \ref{s:experiments}. Similarly to the test data distributions, the hyperparameters of GNN model $m$ (for a particular dataset) is determined by a sample or specification $(h_{m1}, h_{m2}, \ldots) \in \mathcal{H}_m =  H_{m1}\times H_{m2}\times\ldots$. The complete set of inputs to the \gw method are:
\begin{itemize}
    \item \textbf{Models}: list of models $m_1, m_2, \ldots$ and associated hyperparameter spaces $\mathcal{H}_{1}, \mathcal{H}_{2}, \ldots$.
    \item \textbf{Task formulation}: space of possible datasets $\mathcal{D}$, and a probability distribution $\mathcal{P}(\pi_1, \pi_2, \ldots)$ defined on $\mathcal{D}$.
    \item \textbf{Task metric}: test accuracy function $\texttt{EvalMetric}(m, D)$.
    \item \textbf{Size $N$}: number of datasets to sample on the graph world.
\end{itemize}
With these inputs, the \gw system is trivially parallel over $N$ location on the graph world. We give more architecture details in Appendix \ref{s:implementation}. In the following Section \ref{s:experiments}, we specify the above inputs for three different \gw pipelines, and discuss the results of those pipelines in Section \ref{s:insights}.

\subsection{Efficient exploration of graph worlds}\label{ss:gw-exploration}
A key aspect of \gw is the ability to analyze the response of GNN models to the task generator parameters described previously. However, not all configurations $(\pi_1, \pi_2, \ldots)$ in parameter space will provide equivalent insights. As a trivial example, extremely small values of $\pi_i=$ number of vertices (such as 2 or 3) will clearly not be useful to exploring other parameters, like edge density, or the skewness of the degree distribution, since GNN models will either perform poorly or perfectly on trivially-sized graphs regardless of other parameters.

We provide a methodology to mine a large (random) sample of \gw generator configurations for the most ``affective'' configuration, meaning that deviations from that configuration affect GNN model performance most strongly. Assume we have generated a graph world for a given task $T$ with generator parameter space $\Pi$, and at each location $k$ at the graph world we have a sampled configuration $\hat{\Pi}_k\in \Pi$ and an average GNN test metric $z_k$. Conceptually, a sampled configuration $\hat{\Pi} = (\pi_1, \pi_2, \ldots)$ is most affective if for every $\pi_i\in\hat{\Pi}$, changing the value of any other parameter $\pi_j$ produces variance in the test metrics of GNN models.

To find such a configuration, we perform marginal optimization on the space of parameters $\Pi$. Using the samples $\{(\hat{\Pi}_k, z_k)\}$ on an initial run of a \gw pipeline, we find a (locally) optimal setting for each $\pi_i$ in the following manner. We first bin each dimension of $\Pi$ into a fixed number of quantile bins. Then for each quantized value $\pi_i = x$, we compute the average F statistic \citep{snedecor1957statistical} between the other parameter values $\pi_j$ and the test metric $z$ (on graph world locations where $\pi_i = x$). We then set $\pi_i$ to the $x$ value that produced the highest F statistic. This produces an optimal generator configuration from which we can efficiently sample a smaller but still-interesting graph world. In Section \ref{s:experiments} we describe how we apply this technique to \gw experiments.

%% file: contents/4.experiments.tex
\section{Experimental Design}\label{s:experiments}
In this section, we introduce novel experimental design for the \gw method, showing how to efficiently sample a useful part of any graph world. We describe three tasks -- node classification, graph classification, and link prediction -- and how they are generated in various graph worlds. We also list the GNN models tested with the \gw applications, and present novel \gw modes of hyperparameter tuning and inference.

\input{figures/mode2_sbm}

\subsection{Methods}\label{ss:models}
For the experiments in this paper, we choose 11 representative GNNs and 3 baselines to illustrate the strengths of our proposed approach.  
\begin{itemize}[leftmargin=*]
    \item \textbf{ARMA} \cite{bianchi2021graph}: A GNN with auto-regressive moving average filters.
    \item \textbf{APPNP}  \cite{klicpera2018predict}: One of the first `linear' GNNs, accelerating propagation using Personalized PageRank.
    \item \textbf{FiLM} \cite{brockschmidt2020gnn}: A model that modulates an incoming message by the features of the target node.
    \item \textbf{GAT} \cite{velivckovic2017graph}: An early model of graph attention.
    \item \textbf{GATv2} \cite{brody2021attentive}:  An improved variant of graph attention that allows any node to attend to any other one.
    \item \textbf{GCN} \cite{kipf2016semi}: A seminal model that averages neighbor state at each iteration.
    \item \textbf{GIN} \cite{xu2018powerful}: This model uses MLPs to transform summations of neighbor features.
    \item \textbf{GraphSAGE} \cite{hamilton2017inductive}: A variant of the GCN which adds uses sampling and improved propagation of the hidden state.
    \item \textbf{SGC} \cite{wu2019simplifying}: Linear GNNs using matrix multiplication.
    \item \textbf{SuperGAT} \cite{kim2021how}: An approach to improve the graph attention layer.
    \item \textbf{Transformer} \cite{shi2020masked}: A multi-head attention-based model.
\end{itemize}
In addition we examine several baselines which are not GNNs:
\begin{itemize}[leftmargin=*]
    \item \textbf{Linear Regression}: (\textit{graph property prediction only}) Simple ordinary least-squares with edge density as the sole feature.
    \item \textbf{Multi-Layer Perceptron}: (\textit{features only}) Transforms the node features via a DNN  for classification.
    \item \textbf{Personalized PageRank} \cite{brin1998anatomy}: (\textit{graph only}) Predicts node labels for unseen nodes via Personalized PageRank seeded by the labelled nodes. For each unseen node, we compute the total probability mass that comes from the vertices of each label, and pick the label with the highest score.
    \item \textbf{Link prediction heuristics} \cite{marareproducible}: (\textit{graph only}) Creates a link prediction ranking using eight different reweighting schemes for counting common neighbors. We pick one of the following schemes: (i) S{\o}rensen--Dice coefficient \cite{sorensen1948method,dice1945measures}, (ii) cosine similarity, (iii, iv) hub-promoted and hub-suppressed similarity \cite{ravasz2002hierarchical}, (v) Jaccard similarity \cite{jaccard1912distribution}, (vi) Adamic--Adar index \cite{adamic2003friends}, (vii) Resource Allocation index \cite{zhou2009predicting}, and (viii) Leicht--Holme--Newman similarity \cite{leicht2006vertex}.
\end{itemize}
The GNN models use the reference implementations from the PyTorch-Geometric library \cite{Fey/Lenssen/2019}.
We note that by design it is trivial to add additional models into \gw.

\subsection{Hyperparameter Optimization}\label{ss:hyper}
GNN hyperparameter tuning is essential for understanding model performance, and is an aspect of GNN experimentation that can be efficiently explored with \gw. A \gw pipeline can be run in one of three hyperparameter modes:
\begin{enumerate}
    \item[\textbf{Mode 1}] Each model $m$ is trained and tested with a random draw $(h_{m1}, h_{m2}, ...)\in\mathcal{H}_m$, its hyperparameter configuration space.
    \item[\textbf{Mode 2}] Assume a \gw pipeline has already been run in Mode 1. For any model $m$, let $\hat{H}_{i}$ be the $i$-th unique configuration sampled (at any location in the graph world). Let $\mathcal{D}_{i}$ be the collection of \gw datasets for which $\hat{H}_{i}$ was sampled for $m$. Mode 2 is to run another \gw pipeline with the best config $H_m^\ast$ defined as:
    \begin{equation}
        H^\ast = \argmax_{\hat{H}_{i}}|\mathcal{D}_{i}|^{-1}\sum_{D\in\mathcal{D}_{i}}\texttt{EvalMetric}(m(\hat{H}_{i}),D).
    \end{equation}
    Intuitively, we pick the hyperparameters that achieve the best average performance across all \gw samples.
    \item[\textbf{Mode 3}] Each model $m$ receives a budget of $t$ tuning rounds, and the hyperparameter configuration which performed best on a held-out validation set is used to compute the test metric.
\end{enumerate}
We give more model and hyperparameter details in Appendix \ref{ss:models-appendix}.

\subsection{Tasks}
Here we describe three tasks that will be explored with three separate \gw pipelines. We provide the ranges of generator distribution parameters in Appendix \ref{ss:generator-appendix}. In Section \ref{s:insights} we show results from each of tasks run in each of the hyperparameter tuning modes described previously in Section \ref{ss:hyper}.

\input{figures/mode2_lp}

\subsubsection{Node Classification (NC)}
In this \gw experiment, we generate graphs using the Stochastic Block Model (SBM). First, node labels (classification targets) are generated from a multinomial distribution, which define the node clusters. Edges are generated as Bernoulli random variables following within-block probability $p$ and between-block probability $q$ ($p\geq q$). Node features are generated from a within-cluster multivariate Normal distribution, with unit (diagonal) covariance, and cluster centers are drawn from a prior multivariate Normal. The variance of the prior controls the degree of separation between the cluster feature centers. The number of clusters, the cluster size, and the power law of the expected degree distribution, are also varied as described further in the Appendix. We tune and test GNN models with ROC-AUC one-vs-rest (AUC-ovr). We train models on a random sample of 5 nodes per-class, tune on a (disjoint) random sample of 5 nodes per-class, and test on the rest of the nodes.

\subsubsection{Link Prediction (LP)}
In this \gw experiment, we generate graphs using the SBM, as for node classification. However, to simulate a link prediction setting, we randomly split edges into training, validation, and test sets. The task is to predict the "unseen" edges in the test set, which we evaluate with the ROC-AUC metric against randomly chosen negatives, imitating the setting in \cite{grover2016node2vec}. We train on 80\% of the edges, tune on 10\% of the edges, and test on the remaining 10\%.

\subsubsection{Graph Property Prediction (GPP)}
In this \gw experiment, we generate a dataset of small Erd\H{o}s-Renyi random graphs. The task is to infer the number of a certain motif in each test graph. In this paper, we evaluate tailed-triangle motif counting. We evaluate the models with scaled mean-squared-error (S-MSE): $\sum(y_i - \hat{y}_i)^2 / \sum(y_i - \bar{y})^2$, which is comparable across datasets with different scales of motif counts. As in \cite{chen2020can}, we give a dummy unit one-dimensional feature to each node. Here, the number of training graphs is a variable parameter (see Appendix \ref{ss:generator-appendix} for details). We tune on 20\% of the data, and test on the remainder.

\section{Results and Insights}\label{s:insights}
In this section, we present preliminary results from the \gw pipeline. Following the experimental design described in the previous section, we ran nine \gw pipelines, one for each of three tasks, and with all three hyperparameter optimization modes per task (see Section \ref{ss:hyper}). To sample more efficiently from useful regions of the graph worlds, we applied the \gw exploration technique described in Section \ref{ss:gw-exploration} to the Mode 1 experiments, extracting default configurations to use for Mode 2 and Mode 3. In those modes, we sample \emph{only one} parameter from the generator, holding the other parameters fixed at the default config. We provide the default configuration values in Section \ref{ss:generator-appendix}.

The following two sections cover \emph{global}-distribution and \emph{marginal}-distribution insights uncovered by \gw. While our primary aim is not to confirm or overturn established results in the literature, to illustrate the utility of \gw, we point out previously-unseen behavior of GNN models which \gw has exposed with synthetic graph generation.

\subsection{Global Results}\label{ss:global-results}

In this section we cover two insights from analyzing the global performance of GNN models on complete \gw distributions.

\input{tables/ss-mode-avgs}
\input{tables/nc-mode-avgs}
\input{tables/lp-mode-avgs}

\subsubsection{Insight: Globally Optimal Hyperparameters Work Well}
% configurations outperforms local tuning.}
As shown in Tables \ref{tab:ss-mode-avgs}, \ref{tab:nc-mode-avgs}, and \ref{tab:lp-mode-avgs}, most models tested in \gw Mode 2 (training each model with its best-performing hyperparameter set from Mode 1) uniformly outperform their counterparts tested in Mode 1 (with the exception of baselines like PPR, which do not have tunable hyperparameters). While Mode 2 models still do not uniformly outperform Mode 3 models (which each receive a budget of 100 tuning rounds), they do come close, especially on the link prediction task. This shows that \gw can cheaply find hyperparameter improvements that work for a large variety of graphs.

Furthermore, as seen in Appendix Figure \ref{fig:config-performance}, many of the top-performing hyperparameter configurations from \gw Mode 1 have similar average test accuracy (there is no stand-out best configuration). Figure \ref{fig:config-performance} displays the dropoff in performance of all unique sampled hyperparameter configurations for the \gw Link Prediction Mode 1 experiment. Each configuration has an average test metric score, averaged over each graph world location at which it was sampled. The lines in this plot represent the \emph{ordered} scores for each model, the x-axis representing the inverse percentile rank of the score. This depiction illustrates that for most models, there is no "elbow" \cite{thorndike1953belongs} or clear break between top-performing hyperparameter configuration and the next 10--20 top performing configurations.

While we are cautious about how these observations apply to graphs with more complex features, it does suggest that finding good hyper-parameters for models in \gw is low-cost.
As a result of this finding all figures and tables (other than Tables \ref{tab:ss-mode-avgs}--\ref{tab:lp-mode-avgs}) contain data from just \gw Mode 2 experiments, as in Mode 2 we are able to sample more graphs than in Mode 3 with the same computational budget.

Another important observation from \gw mode comparisons regard one family of models---variations of Graph Attention Networks---that exhibit high sensitivity for hyperparameters in the node classification experiments in Table\ref{tab:nc-mode-avgs}. While not achieving the absolute best performance, the difference between Mode 2 and Mode 3 is the most profound for GAT models. We can also compare the improvements to the original GAT architecture. We observe that SuperGAT achieves significantly better performance in Mode 3, whereas GATv2 struggles to improve over its parent model.

\subsubsection{Insight: Most GNNs Can't Count Substructures}
Table \ref{tab:ss-mode-avgs} shows that among GNNs we tested on the Graph Property Prediction graph world, only GIN achieved better-than-mean-fitting MSE, along with simple linear regression with edge density as a feature, which outperformed all other GNNs. In fact, all other methods performed no better (on average) than the naive mean-predictor which produces a scaled MSE of exactly 1.0. This result both accords with and contrasts with various results from the paper ``Can graph neural networks count substructures?" \citep{chen2020can}, which contains a synthetic data experiment that \gw replicates thousands of times. We make the following observations and comparisons:
\begin{itemize}
\item The conclusion of \cite{chen2020can} generally holds that "Message-Passing Neural Networks cannot count substructures".  Interestingly however, the GIN architecture is able to learn a representation which has some utility for the task, significantly outperforming the linear regression baseline. This makes intuitive sense, as GIN was developed specifically to be more-expressive for whole-graph encoding tasks.
\item We show that simple linear regression using edge density as a feature can count substructures better than most GNNs. 
\end{itemize}

With \gw, we reveal a more controlled and reproducible study into substructure counting, using appropriately-scaled MSE, showing that most GNNs fail on the task, but surprisingly GIN does not (even though none of the GNNs are given meaningful features). Due to the lack of good performance of most GNNs on this task, for the rest of this section we analyze \gw results only on Link Prediction and Node Classification tasks.

\subsection{Marginal Results}\label{ss:marginal-results}
We now turn to marginal analysis of GNN models. Marginal parameter analysis is a unique and powerful property of \gw, allowing us to examine the average response of GNN models to \emph{particular, explainable} characteristics of the task. We rely on plots in figures \ref{fig:surfaceplots}, \ref{fig:nc-plots}, and \ref{fig:lp-plots} for these insights. We produced those plots using data from \gw Mode 2, using only samples in each plot from which the corresponding parameter was varied.

\subsubsection{Insight: GNN models switch ranks outside of standard benchmark space.}
To establish our key empirical result, as seen in the three plots inside Figure \ref{fig:surfaceplots}, we project the \gw node classification task distribution space into a 2-D plane measuring each graph's average degree and edge homogeneity, which is the proportion of edges that connect nodes in the same class \cite{zhu2020beyond}. Our first finding is that standard benchmark graphs (shown as black points on the plot) cover only a small region of this graph space that \gw is able to cover via synthetic graph generation. This adds to the strong overall motivation for the \gw method described in Section \ref{s:motivation}, since these statistics should (intuitively) strongly affect graph convolutions.

On the $z$-axis of each plot in Figure \ref{fig:surfaceplots}, we measure the mean reciprocal rank of GCN, APPNP, and FiLM (respectively) against the other 12 models. Our second finding is that---indeed, as expected---GNN models exhibit high ranking variance across this slice of synthetic graph space. We find sharp MRR phase transitions around 0.5 edge homogeneity, and for lower values of average degree. Furthermore, importantly, standard benchmark datasets mostly avoid regions of phase transitions. This strongly suggests that standard benchmark datasets are insufficient to produce generalizable rankings of models and that there is a serious risk of overfitting to the small number of available benchmark datasets for GNNs.
We are hopeful that more comprehensive benchmarking by the means of \gw will help the field to continue to make forward progress.

\subsubsection{Insight: GNNs respond surprisingly to graph characteristics}
Our \gw experiments on node classification (NC) and link prediction (LP) tasks offer both intuitive and counter-intuitive insights about GNN responsiveness to graph and node feature distributions. We make the following observations:
\begin{itemize}[leftmargin=*]
    \item \textbf{Number of vertices doesn't matter.} Across NC and LP tasks, the size of the graph (number of vertices) has negligible effect on test AUC. This suggests that in many cases, it may be sufficient to test new GNN architectures on small graphs produced by \gw, rather than focusing on large graphs that are currently being proposed as a one-size-fits-all solution to benchmarking GNNs \cite{hu2020open}.
    \item \textbf{NC: Differential sensitivity to graph and feature signal.} For NC, most models increased test AUC as the $p$-to-$q$ ratio (graph cluster signal) and feature-center-distance (feature cluster signal) increased. Interestingly, attention-based methods (GAT, GATv2) did not respond as well to these parameters, and seemed to respond negatively to stronger feature signal. FiLM and MLP, which depend strongly on the features, do not respond at all to \nicefrac{p}{\!q}, but are among the top-performers as the feature signal increases.
    \item \textbf{LP: weaker dependence on cluster strength.} For link prediction, interestingly, the distance between feature clusters and \nicefrac{p}{\!q} ratio do not have as strong of an effect on models as in the NC task, and some models even seem to exhibit local-maxima behavior in response to these parameters. This suggests that these models (GAT, GATv2, and FiLM) can not harness very powerful node features and translate them to positional embedding for link prediction.
\end{itemize}

The insights described in this section and Section \ref{ss:global-results} are not possible without a method like \gw, which (1) has the capability to generate millions of test datasets with diverse characteristics and (2) \emph{logs} the dataset characteristics along with test metrics for each model. The insights above are certainly not all that could be gleaned from \gw experiments, or even the particular \gw experiments that we ran. We hope that future GNN researchers will include \gw studies as complements to real data analyses.

\subsection{Cost and scale}\label{s:cost}

It is relatively cheap to perform large-scale GNN model analyses such as those in this paper with \gw. Our node classification experiments featured in the paper cost under \$120, involving 13 models on 1M+ synthetic benchmark datasets, with no GPUs and negligible RAM. By comparison, the experiment with real-world OGB data from \cite{li2021deeperbiggerbetter} involved only 3 models, only 1 Open Graph Benchmark (OGB) dataset, 4 GPUs, and >480GB of RAM per >24 CPUs. These resources would cost >\$500 on modern cloud compute platforms (see \url{https://cloud.google.com/compute}). Additionally, that OGB experiment completed in >40hrs, whereas our GraphWorld experiments took ~10hrs total.

%% file: figures/mode2_sbm.tex
\begin{figure}[!th]
\centering
\begin{tikzpicture}
\begin{groupplot}[group style={
    group name=myplot,
    group size= 2 by 2,
    horizontal sep=0.9cm,
    vertical sep=0.9cm},
    height=4.5cm,
    width=0.57\columnwidth,
    title style={at={(0.5,0.97)},anchor=south},
    every axis x label/.style={at={(axis description cs:0.5,-0.075)}, anchor=north},
    every axis y label/.style={at={(axis description cs:-0.1,0.5)}, anchor=south, rotate=90},
    legend image code/.code={
    \draw
    plot coordinates {
    (0cm,0cm)
    (0.45cm,0cm)         %% default is (0.6cm,0cm)
    };%
    }
    ]
\nextgroupplot[
 	%title = \textbf{Meow},
 	legend columns=4,
	legend style={at={(1.05,1.05)},anchor=south},
	ylabel=ROC AUC$\times100$,
	xlabel=\nicefrac{p}{q} ratio,
	xmin=1,
	xmax=10,
	ymin=50,
	ymax=75,
    ytick={50,60,70,80,90,100},
    xtick={1,2,4,6,8,10},
    legend entries={GCN, GIN, GAT, GraphSAGE, APPNP, SGC, ARMA, GATv2, FiLM, SuperGAT, Tranformer, MLP, Baseline},
]
\addplot[very thick,color=cycle6] table[x=x,y=GCN_mean] {data/kdd_50/mode2_sbm_p_to_q_ratio.csv};
\addplot[very thick,color=cycle2] table[x=x,y=GIN_mean] {data/kdd_50/mode2_sbm_p_to_q_ratio.csv};
\addplot[very thick,color=cycle5] table[x=x,y=GAT_mean] {data/kdd_50/mode2_sbm_p_to_q_ratio.csv};
\addplot[very thick,color=cycle3] table[x=x,y=GraphSAGE_mean] {data/kdd_50/mode2_sbm_p_to_q_ratio.csv};
\addplot[very thick,color=cycle4] table[x=x,y=APPNP_mean] {data/kdd_50/mode2_sbm_p_to_q_ratio.csv};
\addplot[very thick,color=cycle7] table[x=x,y=SGC_mean] {data/kdd_50/mode2_sbm_p_to_q_ratio.csv};
\addplot[very thick,color=cycle11] table[x=x,y=ARMA_mean] {data/kdd_50/mode2_sbm_p_to_q_ratio.csv};
\addplot[very thick,color=cycle8] table[x=x,y=GATv2_mean] {data/kdd_50/mode2_sbm_p_to_q_ratio.csv};
\addplot[very thick,color=cycle10] table[x=x,y=FiLM_mean] {data/kdd_50/mode2_sbm_p_to_q_ratio.csv};
\addplot[very thick,color=cycle12] table[x=x,y=SuperGAT_mean] {data/kdd_50/mode2_sbm_p_to_q_ratio.csv};
\addplot[very thick,color=cycle14] table[x=x,y=Transformer_mean] {data/kdd_50/mode2_sbm_p_to_q_ratio.csv};
\addplot[very thick,color=cycle9] table[x=x,y=MLP_mean] {data/kdd_50/mode2_sbm_p_to_q_ratio.csv};
\addplot[very thick,color=cycle1] table[x=x,y=PPRBaseline_mean] {data/kdd_50/mode2_sbm_p_to_q_ratio.csv};

\nextgroupplot[
 	%title = \textbf{Meow},
	xlabel=feature center distance,
	xmin=0,
	xmax=5,
	ymin=50,
	ymax=100,
    ytick={50,60,70,80,90,100},
    xtick={0,1,2,3,4,5},
]

\addplot[very thick,color=cycle6] table[x=x,y=GCN_mean] {data/kdd_50/mode2_sbm_feature_center_distance.csv};
\addplot[very thick,color=cycle2] table[x=x,y=GIN_mean] {data/kdd_50/mode2_sbm_feature_center_distance.csv};
\addplot[very thick,color=cycle5] table[x=x,y=GAT_mean] {data/kdd_50/mode2_sbm_feature_center_distance.csv};
\addplot[very thick,color=cycle3] table[x=x,y=GraphSAGE_mean] {data/kdd_50/mode2_sbm_feature_center_distance.csv};
\addplot[very thick,color=cycle4] table[x=x,y=APPNP_mean] {data/kdd_50/mode2_sbm_feature_center_distance.csv};
\addplot[very thick,color=cycle7] table[x=x,y=SGC_mean] {data/kdd_50/mode2_sbm_feature_center_distance.csv};
\addplot[very thick,color=cycle11] table[x=x,y=ARMA_mean] {data/kdd_50/mode2_sbm_feature_center_distance.csv};
\addplot[very thick,color=cycle8] table[x=x,y=GATv2_mean] {data/kdd_50/mode2_sbm_feature_center_distance.csv};
\addplot[very thick,color=cycle10] table[x=x,y=FiLM_mean] {data/kdd_50/mode2_sbm_feature_center_distance.csv};
\addplot[very thick,color=cycle12] table[x=x,y=SuperGAT_mean] {data/kdd_50/mode2_sbm_feature_center_distance.csv};
\addplot[very thick,color=cycle14] table[x=x,y=Transformer_mean] {data/kdd_50/mode2_sbm_feature_center_distance.csv};
\addplot[very thick,color=cycle9] table[x=x,y=MLP_mean] {data/kdd_50/mode2_sbm_feature_center_distance.csv};
\addplot[very thick,color=cycle1] table[x=x,y=PPRBaseline_mean] {data/kdd_50/mode2_sbm_feature_center_distance.csv};

\nextgroupplot[
 	%title = \textbf{Meow},
	xlabel=avg.\ degree,
	xmin=1,
	xmax=20,
	ymin=50,
	ymax=80,
	ylabel=ROC AUC$\times100$,
    xtick={1, 5, 10, 15, 20},
]
\addplot[very thick,color=cycle6] table[x=x,y=GCN_mean] {data/kdd_50/mode2_sbm_avg_degree.csv};
\addplot[very thick,color=cycle2] table[x=x,y=GIN_mean] {data/kdd_50/mode2_sbm_avg_degree.csv};
\addplot[very thick,color=cycle5] table[x=x,y=GAT_mean] {data/kdd_50/mode2_sbm_avg_degree.csv};
\addplot[very thick,color=cycle3] table[x=x,y=GraphSAGE_mean] {data/kdd_50/mode2_sbm_avg_degree.csv};
\addplot[very thick,color=cycle4] table[x=x,y=APPNP_mean] {data/kdd_50/mode2_sbm_avg_degree.csv};
\addplot[very thick,color=cycle7] table[x=x,y=SGC_mean] {data/kdd_50/mode2_sbm_avg_degree.csv};
\addplot[very thick,color=cycle11] table[x=x,y=ARMA_mean] {data/kdd_50/mode2_sbm_avg_degree.csv};
\addplot[very thick,color=cycle8] table[x=x,y=GATv2_mean] {data/kdd_50/mode2_sbm_avg_degree.csv};
\addplot[very thick,color=cycle10] table[x=x,y=FiLM_mean] {data/kdd_50/mode2_sbm_avg_degree.csv};
\addplot[very thick,color=cycle12] table[x=x,y=SuperGAT_mean] {data/kdd_50/mode2_sbm_avg_degree.csv};
\addplot[very thick,color=cycle14] table[x=x,y=Transformer_mean] {data/kdd_50/mode2_sbm_avg_degree.csv};
\addplot[very thick,color=cycle9] table[x=x,y=MLP_mean] {data/kdd_50/mode2_sbm_avg_degree.csv};
\addplot[very thick,color=cycle1] table[x=x,y=PPRBaseline_mean] {data/kdd_50/mode2_sbm_avg_degree.csv};

\nextgroupplot[
 	%title = \textbf{Meow},
	xlabel=number of nodes,
	xmin=128,
	xmax=512,
	ymin=50,
	ymax=80,
    xtick={128, 200, 300, 400, 512},
]

\addplot[very thick,color=cycle6] table[x=x,y=GCN_mean] {data/kdd_50/mode2_sbm_nvertex.csv};
\addplot[very thick,color=cycle2] table[x=x,y=GIN_mean] {data/kdd_50/mode2_sbm_nvertex.csv};
\addplot[very thick,color=cycle5] table[x=x,y=GAT_mean] {data/kdd_50/mode2_sbm_nvertex.csv};
\addplot[very thick,color=cycle3] table[x=x,y=GraphSAGE_mean] {data/kdd_50/mode2_sbm_nvertex.csv};
\addplot[very thick,color=cycle4] table[x=x,y=APPNP_mean] {data/kdd_50/mode2_sbm_nvertex.csv};
\addplot[very thick,color=cycle7] table[x=x,y=SGC_mean] {data/kdd_50/mode2_sbm_nvertex.csv};
\addplot[very thick,color=cycle11] table[x=x,y=ARMA_mean] {data/kdd_50/mode2_sbm_nvertex.csv};
\addplot[very thick,color=cycle8] table[x=x,y=GATv2_mean] {data/kdd_50/mode2_sbm_nvertex.csv};
\addplot[very thick,color=cycle10] table[x=x,y=FiLM_mean] {data/kdd_50/mode2_sbm_nvertex.csv};
\addplot[very thick,color=cycle12] table[x=x,y=SuperGAT_mean] {data/kdd_50/mode2_sbm_nvertex.csv};
\addplot[very thick,color=cycle14] table[x=x,y=Transformer_mean] {data/kdd_50/mode2_sbm_nvertex.csv};
\addplot[very thick,color=cycle9] table[x=x,y=MLP_mean] {data/kdd_50/mode2_sbm_nvertex.csv};
\addplot[very thick,color=cycle1] table[x=x,y=PPRBaseline_mean] {data/kdd_50/mode2_sbm_nvertex.csv};
\end{groupplot}
\end{tikzpicture}
\vspace*{-5mm}
\caption{\gw node classification results (mode 2).}\label{fig:nc-plots}
\end{figure}
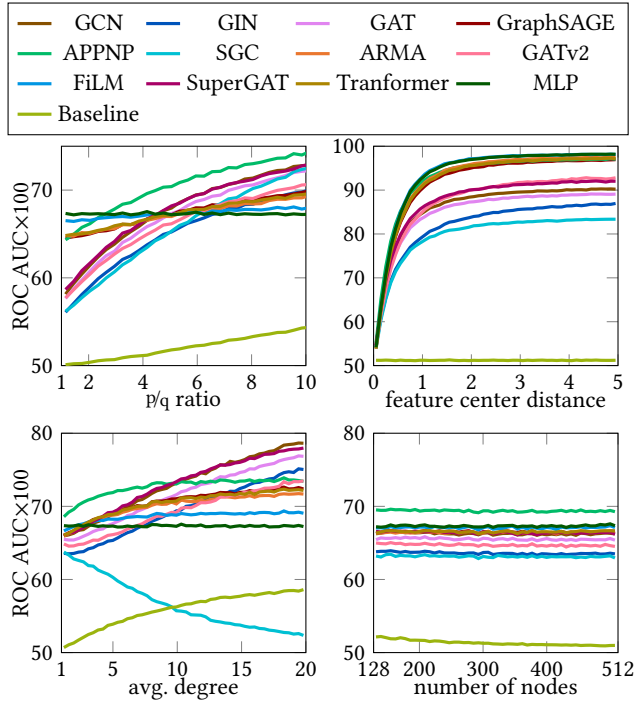

%% file: figures/mode2_lp.tex
\begin{figure}[!th]
\centering
\begin{tikzpicture}
\begin{groupplot}[group style={
    group name=myplot,
    group size= 2 by 2,
    horizontal sep=0.9cm,
    vertical sep=0.9cm},
    height=4.5cm,
    width=0.57\columnwidth,
    title style={at={(0.5,0.97)},anchor=south},
    every axis x label/.style={at={(axis description cs:0.5,-0.075)}, anchor=north},
    every axis y label/.style={at={(axis description cs:-0.1,0.5)}, anchor=south, rotate=90},
    legend image code/.code={
    \draw
    plot coordinates {
    (0cm,0cm)
    (0.45cm,0cm)         %% default is (0.6cm,0cm)
    };%
    }
    ]
\nextgroupplot[
 	%title = \textbf{Meow},
 	legend columns=4,
	legend style={at={(1.05,1.05)},anchor=south},
	ylabel=ROC AUC$\times100$,
	xlabel=\nicefrac{p}{q} ratio,
	xmin=1,
	xmax=10,
	ymin=50,
	ymax=86,
    ytick={60,70,80},
    xtick={1,2,4,6,8,10},
    legend entries={GCN, GIN, GAT, GraphSAGE, APPNP, SGC, ARMA, GATv2, FiLM, SuperGAT, Tranformer, MLP, Baseline},
]
\addplot[very thick,color=cycle6] table[x=x,y=GCN_mean] {data/kdd_50/mode2_lp_p_to_q_ratio.csv};
\addplot[very thick,color=cycle2] table[x=x,y=GIN_mean] {data/kdd_50/mode2_lp_p_to_q_ratio.csv};
\addplot[very thick,color=cycle5] table[x=x,y=GAT_mean] {data/kdd_50/mode2_lp_p_to_q_ratio.csv};
\addplot[very thick,color=cycle3] table[x=x,y=GraphSAGE_mean] {data/kdd_50/mode2_lp_p_to_q_ratio.csv};
\addplot[very thick,color=cycle4] table[x=x,y=APPNP_mean] {data/kdd_50/mode2_lp_p_to_q_ratio.csv};
\addplot[very thick,color=cycle7] table[x=x,y=SGC_mean] {data/kdd_50/mode2_lp_p_to_q_ratio.csv};
\addplot[very thick,color=cycle11] table[x=x,y=ARMA_mean] {data/kdd_50/mode2_lp_p_to_q_ratio.csv};
\addplot[very thick,color=cycle8] table[x=x,y=GATv2_mean] {data/kdd_50/mode2_lp_p_to_q_ratio.csv};
\addplot[very thick,color=cycle10] table[x=x,y=FiLM_mean] {data/kdd_50/mode2_lp_p_to_q_ratio.csv};
\addplot[very thick,color=cycle12] table[x=x,y=SuperGAT_mean] {data/kdd_50/mode2_lp_p_to_q_ratio.csv};
\addplot[very thick,color=cycle14] table[x=x,y=Transformer_mean] {data/kdd_50/mode2_lp_p_to_q_ratio.csv};
\addplot[very thick,color=cycle9] table[x=x,y=MLP_mean] {data/kdd_50/mode2_lp_p_to_q_ratio.csv};
\addplot[very thick,color=cycle1] table[x=x,y=Baseline_mean] {data/kdd_50/mode2_lp_p_to_q_ratio.csv};

\nextgroupplot[
 	%title = \textbf{Meow},
	xlabel=feature center distance,
	xmin=0,
	xmax=5,
	ymin=50,
	ymax=85,
    xtick={0,1,2,3,4,5},
]

\addplot[very thick,color=cycle6] table[x=x,y=GCN_mean] {data/kdd_50/mode2_lp_feature_center_distance.csv};
\addplot[very thick,color=cycle2] table[x=x,y=GIN_mean] {data/kdd_50/mode2_lp_feature_center_distance.csv};
\addplot[very thick,color=cycle5] table[x=x,y=GAT_mean] {data/kdd_50/mode2_lp_feature_center_distance.csv};
\addplot[very thick,color=cycle3] table[x=x,y=GraphSAGE_mean] {data/kdd_50/mode2_lp_feature_center_distance.csv};
\addplot[very thick,color=cycle4] table[x=x,y=APPNP_mean] {data/kdd_50/mode2_lp_feature_center_distance.csv};
\addplot[very thick,color=cycle7] table[x=x,y=SGC_mean] {data/kdd_50/mode2_lp_feature_center_distance.csv};
\addplot[very thick,color=cycle11] table[x=x,y=ARMA_mean] {data/kdd_50/mode2_lp_feature_center_distance.csv};
\addplot[very thick,color=cycle8] table[x=x,y=GATv2_mean] {data/kdd_50/mode2_lp_feature_center_distance.csv};
\addplot[very thick,color=cycle10] table[x=x,y=FiLM_mean] {data/kdd_50/mode2_lp_feature_center_distance.csv};
\addplot[very thick,color=cycle12] table[x=x,y=SuperGAT_mean] {data/kdd_50/mode2_lp_feature_center_distance.csv};
\addplot[very thick,color=cycle14] table[x=x,y=Transformer_mean] {data/kdd_50/mode2_lp_feature_center_distance.csv};
\addplot[very thick,color=cycle9] table[x=x,y=MLP_mean] {data/kdd_50/mode2_lp_feature_center_distance.csv};
\addplot[very thick,color=cycle1] table[x=x,y=Baseline_mean] {data/kdd_50/mode2_lp_feature_center_distance.csv};

\nextgroupplot[
 	%title = \textbf{Meow},
	xlabel=avg.\ degree,
	xmin=1,
	xmax=20,
	ylabel=ROC AUC$\times100$,	
	ymax=85,
    xtick={1, 5, 10, 15, 20},
]
\addplot[very thick,color=cycle6] table[x=x,y=GCN_mean] {data/kdd_50/mode2_lp_avg_degree.csv};
\addplot[very thick,color=cycle2] table[x=x,y=GIN_mean] {data/kdd_50/mode2_lp_avg_degree.csv};
\addplot[very thick,color=cycle5] table[x=x,y=GAT_mean] {data/kdd_50/mode2_lp_avg_degree.csv};
\addplot[very thick,color=cycle3] table[x=x,y=GraphSAGE_mean] {data/kdd_50/mode2_lp_avg_degree.csv};
\addplot[very thick,color=cycle4] table[x=x,y=APPNP_mean] {data/kdd_50/mode2_lp_avg_degree.csv};
\addplot[very thick,color=cycle7] table[x=x,y=SGC_mean] {data/kdd_50/mode2_lp_avg_degree.csv};
\addplot[very thick,color=cycle11] table[x=x,y=ARMA_mean] {data/kdd_50/mode2_lp_avg_degree.csv};
\addplot[very thick,color=cycle8] table[x=x,y=GATv2_mean] {data/kdd_50/mode2_lp_avg_degree.csv};
\addplot[very thick,color=cycle10] table[x=x,y=FiLM_mean] {data/kdd_50/mode2_lp_avg_degree.csv};
\addplot[very thick,color=cycle12] table[x=x,y=SuperGAT_mean] {data/kdd_50/mode2_lp_avg_degree.csv};
\addplot[very thick,color=cycle14] table[x=x,y=Transformer_mean] {data/kdd_50/mode2_lp_avg_degree.csv};
\addplot[very thick,color=cycle9] table[x=x,y=MLP_mean] {data/kdd_50/mode2_lp_avg_degree.csv};
\addplot[very thick,color=cycle1] table[x=x,y=Baseline_mean] {data/kdd_50/mode2_lp_avg_degree.csv};

\nextgroupplot[
 	%title = \textbf{Meow},
	xlabel=number of nodes,
	xmin=128,
	xmax=512,
	ymin=60,
	ymax=82,
    xtick={128, 200, 300, 400, 500},
]

\addplot[very thick,color=cycle6] table[x=x,y=GCN_mean] {data/kdd_50/mode2_lp_nvertex.csv};
\addplot[very thick,color=cycle2] table[x=x,y=GIN_mean] {data/kdd_50/mode2_lp_nvertex.csv};
\addplot[very thick,color=cycle5] table[x=x,y=GAT_mean] {data/kdd_50/mode2_lp_nvertex.csv};
\addplot[very thick,color=cycle3] table[x=x,y=GraphSAGE_mean] {data/kdd_50/mode2_lp_nvertex.csv};
\addplot[very thick,color=cycle4] table[x=x,y=APPNP_mean] {data/kdd_50/mode2_lp_nvertex.csv};
\addplot[very thick,color=cycle7] table[x=x,y=SGC_mean] {data/kdd_50/mode2_lp_nvertex.csv};
\addplot[very thick,color=cycle11] table[x=x,y=ARMA_mean] {data/kdd_50/mode2_lp_nvertex.csv};
\addplot[very thick,color=cycle8] table[x=x,y=GATv2_mean] {data/kdd_50/mode2_lp_nvertex.csv};
\addplot[very thick,color=cycle10] table[x=x,y=FiLM_mean] {data/kdd_50/mode2_lp_nvertex.csv};
\addplot[very thick,color=cycle12] table[x=x,y=SuperGAT_mean] {data/kdd_50/mode2_lp_nvertex.csv};
\addplot[very thick,color=cycle14] table[x=x,y=Transformer_mean] {data/kdd_50/mode2_lp_nvertex.csv};
\addplot[very thick,color=cycle9] table[x=x,y=MLP_mean] {data/kdd_50/mode2_lp_nvertex.csv};
\addplot[very thick,color=cycle1] table[x=x,y=Baseline_mean] {data/kdd_50/mode2_lp_nvertex.csv};

\end{groupplot}
\end{tikzpicture}
\vspace*{-5mm}
\caption{\gw link prediction results (mode 2).}\label{fig:lp-plots}
\vspace*{-1mm}
\end{figure}
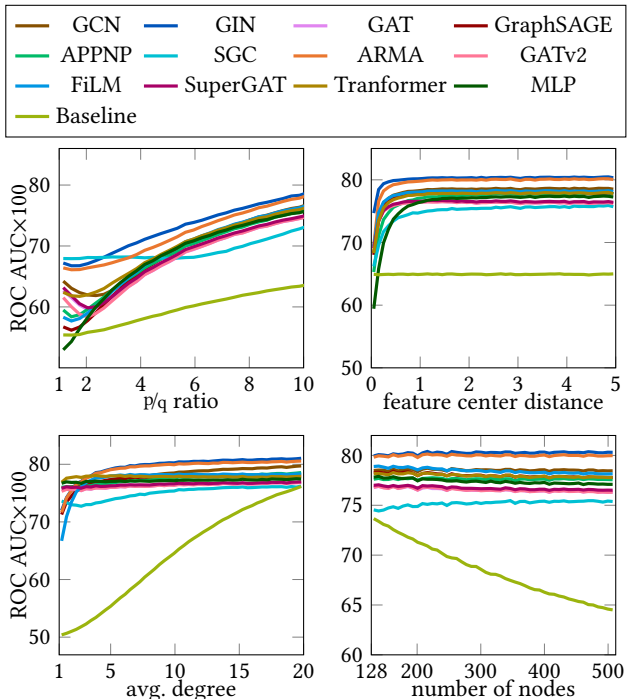

%% file: tables/ss-mode-avgs.tex
\begin{table}[!tb]
\newcolumntype{C}{>{\centering\arraybackslash}X}
\newcolumntype{R}{>{\raggedright\arraybackslash}X}
\small
\begin{tabularx}{\linewidth}{RCCC}
\toprule
\emph{model} & \mbox{Mode 1} & \mbox{Mode 2}& \mbox{Mode 3}\\ 
\cmidrule(lr){1-1}\cmidrule(lr){2-4}
APPNP & $ 1.04 \pm 0.00 $ & $ 1.02 \pm 0.00 $ & $ 1.03 \pm 0.00 $ \\ 
ARMA & $ {1.03 \pm 0.00} $ & $ {0.97 \pm 0.00} $ & $ {0.92 \pm 0.01} $ \\ 
FiLM & $ 1.06 \pm 0.01 $ & $ 1.01 \pm 0.00 $ & $ 1.01 \pm 0.00 $ \\ 
GAT & $ 1.04 \pm 0.00 $ & $ 1.02 \pm 0.00 $ & $ 1.03 \pm 0.00 $ \\ 
GATv2 & $ 1.04 \pm 0.00 $ & $ 1.02 \pm 0.00 $ & $ 1.03 \pm 0.00 $ \\ 
GCN & $ 1.04 \pm 0.00 $ & $ 1.02 \pm 0.00 $ & $ 1.03 \pm 0.00 $ \\ 
GIN & $ \mathbf{0.86 \pm 0.07} $ & $ \mathbf{0.21 \pm 0.01} $ & $ \mathbf{0.33 \pm 0.03} $\\ 
GraphSAGE & $ 1.04 \pm 0.00 $ & $ 1.01 \pm 0.00 $ & $ 1.01 \pm 0.00 $ \\ 
LR & $ \mathbf{0.52 \pm 0.00} $ & $ \mathbf{0.52 \pm 0.00} $ & $ \mathbf{0.51 \pm 0.03} $ \\ 
MLP & $ 1.04 \pm 0.00 $ & $ 1.02 \pm 0.00 $ & $ 1.03 \pm 0.00 $ \\ 
SGC & $ 1.08 \pm 0.00 $ & $ 1.02 \pm 0.00 $ & $ 1.03 \pm 0.01 $ \\ 
SuperGAT & $ 1.04 \pm 0.00 $ & $ 1.02 \pm 0.00 $ & $ 1.03 \pm 0.01 $ \\ 
Transformer & $ 1.04 \pm 0.00 $ & $ 1.00 \pm 0.00 $ & $ 1.00 \pm 0.01 $ \\ 
\bottomrule
\end{tabularx}
\caption{\label{tab:ss-mode-avgs}Graph property prediction performance averages in terms of scaled MSE. Lower is better.}
\vspace{-6mm}
\end{table}

%% file: tables/nc-mode-avgs.tex
\begin{table}[!tb]
\newcolumntype{C}{>{\centering\arraybackslash}X}
\newcolumntype{R}{>{\raggedright\arraybackslash}X}
\small
\begin{tabularx}{\linewidth}{RCCC}
\toprule
\emph{model} & \mbox{Mode 1} & \mbox{Mode 2}& \mbox{Mode 3}\\ 
\cmidrule(lr){1-1}\cmidrule(lr){2-4}
APPNP & \mbox { $ 62.31 \pm 0.05 $ } & \mbox { $ \mathbf{74.16 \pm 0.10} $ } & \mbox { $ {79.16 \pm 0.33} $ } \\ 
ARMA & \mbox { $ \mathbf{65.03 \pm 0.05} $ } & \mbox { $ {71.43 \pm 0.10} $ } & \mbox { $ {77.97 \pm 0.33} $ } \\ 
FiLM & \mbox { $ \mathbf{66.07 \pm 0.05} $ } & \mbox { $ 71.52 \pm 0.11 $ } & \mbox { $ 72.00 \pm 0.34 $ } \\ 
GAT & \mbox { $ 63.38 \pm 0.05 $ } & \mbox { $ 70.42 \pm 0.09 $ } & \mbox { $ \mathbf{80.32 \pm 0.36} $ } \\ 
GATv2 & \mbox { $ 63.40 \pm 0.05 $ } & \mbox { $ 69.79 \pm 0.09 $ } & \mbox { $ \mathbf{80.24 \pm 0.36} $ } \\ 
GCN & \mbox { $ 63.50 \pm 0.05 $ } & \mbox { $ 71.31 \pm 0.09 $ } & \mbox { $ \mathbf{81.46 \pm 0.36} $ } \\ 
GIN & \mbox { $ 61.29 \pm 0.04 $ } & \mbox { $ 68.18 \pm 0.08 $ } & \mbox { $ 76.05 \pm 0.36 $ } \\ 
GraphSAGE & \mbox { $ \mathbf{64.58 \pm 0.05} $ } & \mbox { $ {71.55 \pm 0.10} $ } & \mbox { $ {78.41 \pm 0.33} $ } \\ 
MLP & \mbox { $ {64.14 \pm 0.05} $ } & \mbox { $ 71.46 \pm 0.11 $ } & \mbox { $ 70.92 \pm 0.34 $ } \\ 
PPR & \mbox { $ 59.62 \pm 0.03 $ } & \mbox { $ 52.28 \pm 0.03 $ } & \mbox { $ 59.59 \pm 0.28 $ } \\ 
SGC & \mbox { $ 58.36 \pm 0.04 $ } & \mbox { $ 66.06 \pm 0.09 $ } & \mbox { $ 71.79 \pm 0.42 $ } \\ 
SuperGAT & \mbox { $ 63.58 \pm 0.05 $ } & \mbox { $ 71.56 \pm 0.09 $ } & \mbox { $ \mathbf{81.26 \pm 0.35} $ } \\ 
Transformer & \mbox { $ 64.07 \pm 0.05 $ } & \mbox { $ 71.52 \pm 0.10 $ } & \mbox { $ 77.67 \pm 0.33 $ } \\ 
\bottomrule
\end{tabularx}
\caption{\label{tab:nc-mode-avgs} Node classification performance averages in terms of ROC-AUC. Higher is better.}
\vspace{-6mm}
\end{table}

%% file: tables/lp-mode-avgs.tex
\begin{table}[!tb]
\newcolumntype{C}{>{\centering\arraybackslash}X}
\newcolumntype{R}{>{\raggedright\arraybackslash}X}
\small
\begin{tabularx}{\linewidth}{RCCC}
\toprule
\emph{model} & \mbox{Mode 1} & \mbox{Mode 2}& \mbox{Mode 3} \\ 
\cmidrule(lr){1-1}\cmidrule(lr){2-4}
APPNP & \mbox { $ \mathbf{70.41 \pm 0.02} $ } & \mbox { $ 76.48 \pm 0.01 $ } & \mbox { $ 76.80 \pm 0.11 $ }\\ 
ARMA & \mbox { $ 69.21 \pm 0.03 $ } & \mbox { $ \mathbf{79.49 \pm 0.01} $ } & \mbox { $ \mathbf{79.77 \pm 0.13} $ } \\ 
Heuristics & \mbox { $ 69.65 \pm 0.03 $ } & \mbox { $ 65.92 \pm 0.02 $ } & \mbox { $ 65.91 \pm 0.19 $ }\\ 
FiLM & \mbox { $ 66.28 \pm 0.03 $ } & \mbox { $ 77.40 \pm 0.01 $ } & \mbox { $ 77.22 \pm 0.14 $ } \\ 
GAT & \mbox { $ 60.40 \pm 0.03 $ } & \mbox { $ 75.45 \pm 0.01 $ } & \mbox { $ 75.48 \pm 0.13 $ } \\ 
GATv2 & \mbox { $ 59.62 \pm 0.03 $ } & \mbox { $ 75.44 \pm 0.01 $ } & \mbox { $ 75.50 \pm 0.13 $ } \\ 
GCN & \mbox { $ 64.13 \pm 0.03 $ } & \mbox { $ 77.54 \pm 0.01 $ } & \mbox { $ {78.00 \pm 0.12} $ } \\ 
GIN & \mbox { $ \mathbf{71.06 \pm 0.03} $ } & \mbox { $ \mathbf{79.97 \pm 0.01} $ } & \mbox { $ \mathbf{79.79 \pm 0.13} $ } \\ 
GraphSAGE & \mbox { $ 61.89 \pm 0.04 $ } & \mbox { $ 76.94 \pm 0.01 $ } & \mbox { $ 75.20 \pm 0.14 $ }\\ 
MLP & \mbox { $ 56.28 \pm 0.02 $ } & \mbox { $ 76.01 \pm 0.01 $ } & \mbox { $ 75.90 \pm 0.14 $ } \\ 
SGC & \mbox { $ 66.56 \pm 0.03 $ } & \mbox { $ {74.62 \pm 0.01} $ } & \mbox { $ 75.76 \pm 0.11 $ }\\ 
SuperGAT & \mbox { $ 60.41 \pm 0.03 $ } & \mbox { $ {75.66 \pm 0.01} $ } & \mbox { $ 75.54 \pm 0.13 $ } \\ 
Transformer & \mbox { $ 65.37 \pm 0.03 $ } & \mbox { $ {77.21 \pm 0.01} $ } & \mbox { $ 77.10 \pm 0.14 $ } \\ 
\bottomrule
\end{tabularx}
\caption{\label{tab:lp-mode-avgs}Link prediction mode averages in terms of ROC-AUC.  Higher is better.}
\vspace{-10mm}
\end{table}

%% file: contents/5.conclusion.tex
\vspace{-1.5mm}
\section{Conclusions and Future Work}\label{s:discussion}
The reliance on standardized real-world benchmark datasets in GNN research has been an important way to track technological progress over time on tangible applications. However, as we argued in Section 2, existing benchmark graph datasets are too limited to fully explore the relative performances of GNN models, and can be too costly for the majority of labs to rapidly experiment with. In this paper, we have shown that \gw addresses these problems via the following features:
\begin{enumerate}[leftmargin=*]
\item \textbf{Generalizable analyses.} With tunable parameters for GNN dataset generators, \gw can simulate graphs with far-wider ranges of graph properties than currently exist in any collection of benchmark datasets. As shown in our results (especially Figure \ref{fig:surfaceplots}), the marginal analysis of these parameters and statistics can generate insights about GNN architectures that are unavailable from any small collection of re-used benchmark datasets.

\item \textbf{Reproducibility without overfitting.} Using our code and platform, a researcher can use \gw to test their model as easily as with any existing collection of GNN benchmarks, but without the risk of overfitting to graphs with a limited set of properties.

\item \textbf{Accessibility.} As described in Section \ref{s:cost}, \gw does not require excessive resources, and can actually test many more models at a time for lower cost than standard benchmarks. Furthermore, our experiments show that assessing GNN test performance does not depend on having natural, society-scale graph data. Combining these observations, we have shown that with \gw it is possible to derive new insights with less resources. This is particularly important to facilitate GNN research in smaller labs.
\end{enumerate}

These characteristics make \gw the perfect complement to GNN experiments on graph datasets sourced from nature. While performance on such natural datasets will always be of scientific interest and essential for new research, \gw can expose when progress on them may not transfer to other datasets. More importantly, \gw can help uncover certain distributions of graphs that have not yet been used to test GNNs, which we hope will inspire new architecture development. At Google, we are integrating \gw with GNN experimental pipelines and unit tests, as well as with TF-GNN \cite{tfgnn}.

A complementary and follow-up line of research to our work could be the development of new random graph models with tunable properties that target classes of GNN architectures. For instance, we may wish to design clustered graph models with tunable numbers of certain graph motifs, or attributed graph models with non-trivial feature correlations. \gw is the perfect tool to understand how these variations in these graph properties cause differential responses from various GNN architectures.

Overall, \gw facilitates cheap, comprehensive, and principled investigation into the nuances of GNN model performance. By releasing our codebase at \url{https://github.com/google-research/graphworld} and integrating \gw with TF-GNN, we hope to make GNN experimental results more robust and transferable -- helping researchers reach more reliable conclusions when developing new architectures.

%% file: contents/999.appendix.tex
\section{Implementation and Cost}\label{s:implementation}
Design goals for \gw focused on accessibility, scalability and efficiency; any researcher should be able to run \gw simulations with minimal setup, while having the system automatically scale up experiments to available resources only as needed. To this end, \gw is implemented as a containerized Apache Beam\footnote{\href{https://beam.apache.org/}{https://beam.apache.org/}} pipeline allowing researchers to run a hermetic copy of Graph World on any infrastructure i.e., a local machine, compute cluster, or cloud framework. Experiments in this paper were run on \href{https://cloud.google.com}{Google Cloud Platform (GCP)} using \href{cloud.google.com/dataflow}{Cloud Dataflow}. Experiments were allowed to scale up to a maximum of 1000 workers using \href{https://cloud.google.com/compute/docs/machine-types}{n1-standard-1} machines capable of sampling millions of graphs in $\leq10$ hours. 

\begin{figure*}[h]
  \includegraphics[width=\textwidth]{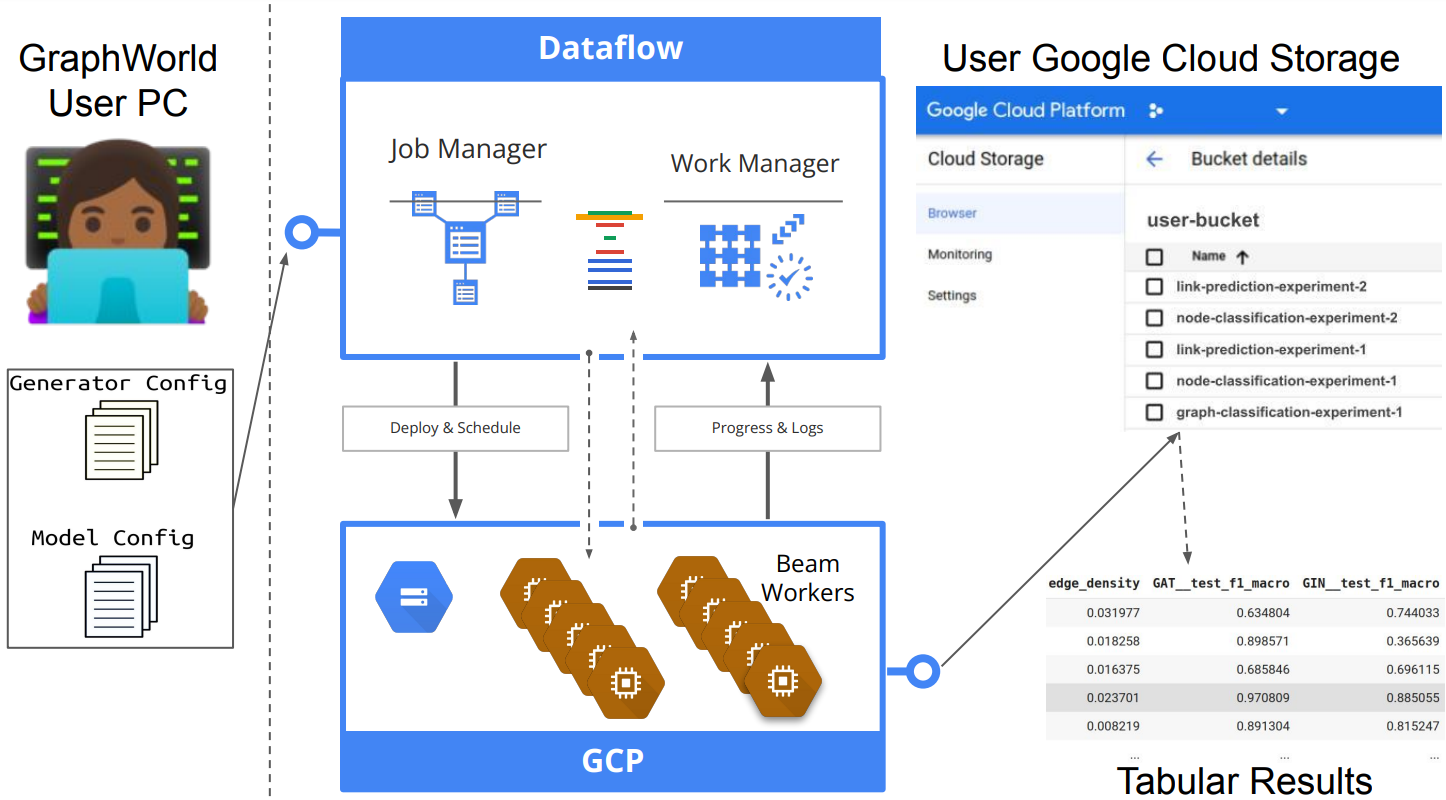}
  \caption{\label{fig:gw-pipeline} On their PC, a \gw user specifies dataset generator and GNN model configurations, and then launches a \gw pipeline, which is submitted on a remote manager. The records of GNN tests are written to Google Cloud Storage (GCS), which we then accumulate into a data table via the GCS API.}
\end{figure*}

Figure \ref{fig:gw-pipeline} shows the design of the \gw distributed processing system. Table \ref{tab:vcpu} shows the number of virtual-CPU hours needed to complete the nine pipelines discussed in Section \ref{s:experiments}.

\section{Experiment Details}
In this Appendix section, we provide more details about \gw pipelines, the task dataset generators, and the model architectures. In particular, we specify all the \gw configuration elements (see Section \ref{s:gw}) of each pipeline described in Section \ref{s:experiments}.

\input{tables/cpu_hours_table}
\input{tables/hyperparam-values}
\input{figures/auc_tuning}

\subsection{Models}\label{ss:models-appendix}
Here we list hyperparameter values available to each GNN model (and some non-GNN models) for tuning. We note that \gw experiments are focused on a comparison of the convolution layers introduced by each model (defined by its corresponding convolution implementation in PyTorch-geometric).

\textbf{Graph Property Prediction.} In order to generate the global readout of the node state, we take the final layer's activations for all the nodes and apply mean pooling to create a graph representation.
This representation is then used for regressing substructure counts.

\subsection{Generator parameters}\label{ss:generator-appendix}
In Tables \ref{tab:sbm-values} and \ref{tab:ss-values}, we list generator parameters for task dataset generators. Table \ref{tab:sbm-values} has parameter values for the Node Classification and Link Prediction pipelines. Table \ref{tab:ss-values} has parameter values for the Graph Property Prediction pipelines. Each table contains the parameter names, their description, and their default values found using the technique described in Section \ref{ss:gw-exploration}.

\input{tables/nc-lp-defaule-vars}
\input{tables/ss-default-vars}

%% file: tables/cpu_hours_table.tex
\begin{table}[!tb]
\newcolumntype{S}{>{\hsize=0.7\hsize}X}
\newcolumntype{Z}{>{\hsize=0.3\hsize}X}
\begin{tabularx}{\linewidth}{ZZSXS}
\toprule
Task & Mode & Samples ($N$) & \mbox{Tuning Rounds} & vCPU hours\\
\midrule
LP & 1 & \num{1e6} & 0 & 1,681 \\
LP & 2 & \num{7e5} & 0 & 1,672 \\
%LP & 3 & \num{7e4} & 10 & 2,017 \\
LP & 3 & \num{7e3} & 100 & 1,896 \\
\cmidrule(lr){1-1}\cmidrule(lr){2-5}
NC & 1 & \num{1e6} & 0 & 1,047 \\
NC & 2 & \num{7e5} & 0 & 755 \\
%NC & 3 & \num{7e4} & 10 & 987 \\
NC & 3 & \num{7e3} & 100 & 937 \\
\cmidrule(lr){1-1}\cmidrule(lr){2-5}
GPP & 1 & \num{1e6} & 0 & 9,767 \\
GPP & 2 & \num{4e5} & 0 & 3,399 \\
%GPP & 3 & \num{4e4} & 10 & 4,020 \\
GPP & 3 & \num{4e3} & 100 & 3,553 \\
\bottomrule
\end{tabularx}
\caption{\label{tab:vcpu} Resource complexity for \gw experiments.}
\end{table}

%% file: tables/hyperparam-values.tex
\begin{table}[!tb]
\newcolumntype{M}{>{\raggedleft\arraybackslash}X}
\begin{tabularx}{\linewidth}{XM@{}}
\toprule
Hyperparameter     &  Values\\
\midrule
Learning Rate     & $[0.01, 0.001, 0.0001]$ \\
Hidden Channels & $[4, 8, 16]$ \\
Number of Layers & $[1, 2, 3, 4]$ \\
Dropout & $[0, 0.3, 0.5, 0.8]$ \\
\mbox{$\alpha$ (APPNP, SGC, and PPR baseline)} & $[0.1, 0.2, 0.3]$ \\
Iterations (APPNP and SGC) & $[5, 10, 15]$ \\
\mbox{\# of attention heads (GATs and Transformer)} & $[1, 2, 3, 4]$ \\
\bottomrule
\end{tabularx}
\caption{\label{tab:hyper-values} Hyperparameter values for all models used by all \gw experiments.}
\end{table}
%\vspace{-5mm}

%% file: figures/auc_tuning.tex
\begin{figure}[!t]
\centering
\begin{tikzpicture}
\begin{axis}[
    height=4cm,
    width=\linewidth,
    legend image code/.code={
    \draw
    plot coordinates {
    (0cm,0cm)
    (0.39cm,0cm)         %% default is (0.6cm,0cm)
    };%
    },
    ylabel style={align=center},
	ylabel={Link prediction \\ ROC AUC$\times100$},
	xmin=1,
	xmax=100,
	ymin=50,
	ymax=100,
 	legend columns=3,
    legend entries={GCN, GIN, GAT, GraphSAGE, APPNP, SGC, ARMA, GATv2, FiLM, SuperGAT, Transformer, MLP, Heuristics},
    xlabel=\% rank of hyperparameter config,
	legend style={at={(1,1.05)},anchor=south east},
    xtick={1,20,40,60,80,100},
    ]
\addplot[very thick,color=cycle6] table[x=x,y=GCN] {data/lp_aucs_tuning.tex};
\addplot[very thick,color=cycle2] table[x=x,y=GIN] {data/lp_aucs_tuning.tex};
\addplot[very thick,color=cycle5] table[x=x,y=GAT] {data/lp_aucs_tuning.tex};
\addplot[very thick,color=cycle3] table[x=x,y=GraphSAGE] {data/lp_aucs_tuning.tex};
\addplot[very thick,color=cycle4] table[x=x,y=APPNP] {data/lp_aucs_tuning.tex};
\addplot[very thick,color=cycle7] table[x=x,y=SGC] {data/lp_aucs_tuning.tex};
\addplot[very thick,color=cycle11] table[x=x,y=ARMA] {data/lp_aucs_tuning.tex};
\addplot[very thick,color=cycle8] table[x=x,y=GATv2] {data/lp_aucs_tuning.tex};
\addplot[very thick,color=cycle10] table[x=x,y=FiLM] {data/lp_aucs_tuning.tex};
\addplot[very thick,color=cycle12] table[x=x,y=SuperGAT] {data/lp_aucs_tuning.tex};
\addplot[very thick,color=cycle14] table[x=x,y=Transformer] {data/lp_aucs_tuning.tex};
\addplot[very thick,color=cycle9] table[x=x,y=MLP] {data/lp_aucs_tuning.tex};
\addplot[very thick,color=cycle1] table[x=x,y=Baseline] {data/lp_aucs_tuning.tex};
\end{axis}
\end{tikzpicture}
\vspace*{-2mm}
\caption{\label{fig:config-performance} Each line is the accuracy dropoff of hyperparameter configurations for a particular model, from the \gw Link Prediction Mode 1 experiment. The $x$-axis is the inverse percentile rank of the hyperparameter configuration, and the $y$-axis is the accuracy. There is no "elbow" \cite{thorndike1953belongs} or clear break between top-performing hyperparameter configuration and the next 10--20 top performing configurations. See Section 5.1 for context.}
\end{figure}

%% file: tables/nc-lp-defaule-vars.tex
\begin{table*}[!th]
\newcolumntype{C}{>{\centering\arraybackslash}X}
\newcolumntype{S}{>{\hsize=0.3\hsize}X}
\newcolumntype{Y}{>{\hsize=0.125\hsize}X}
\newcolumntype{Z}{>{\hsize=0.1\hsize}X}
\begin{tabularx}{\linewidth}{SXYZZ}
\toprule
Parameter Name    &  Description & Values & \mbox{NC} & \mbox{LP}\\
\cmidrule(lr){1-1}\cmidrule(lr){2-5}
nvertex & Number of vertices in the graph. & $[128, 512]$ & 343 & 481 \\
\nicefrac{p}{\!q} ratio & the ratio of in-cluster edge probability to out-cluster edge probability & $[1.0, 10.0]$ & 3.98 & 12.40 \\
avg. degree & the average expected degrees of the nodes & $[1.0, 20.0]$ & 1.75 & 10.12 \\
\mbox{feature center distance} & \mbox{the variance of feature cluster centers, generated from a multivariate Normal} & [0.0, 5.0] & 0.20 & 2.20 \\
num clusters & the number of unique node labels & [2, 6] & 5 & 4\\
cluster size slope & the slope of cluster sizes when index-ordered by size & [0.0, 0.5] & 0.08 & 0.92\\
power exponent & \mbox{the value of the power law exponent used to generate expected node degrees} & [0.5, 1.0] & 1.0 & 1.0\\
\bottomrule
\end{tabularx}
\caption{\label{tab:sbm-values} Node Classification and Link Prediction default generator values used in \gw Mode 2 and Mode 3 experiments.}
\vspace{-4mm}
\end{table*}

%% file: tables/ss-default-vars.tex
\begin{table*}[!th]
\newcolumntype{C}{>{\centering\arraybackslash}X}
\newcolumntype{S}{>{\hsize=0.5\hsize}X}
\begin{tabularx}{\linewidth}{SXSS}
\toprule
Parameter Name    &  Description & Values & Default \\
\cmidrule(lr){1-1}\cmidrule(lr){2-4}
ngraphs & Number of graphs in each dataset. & $[100, 500]$ & 499 \\
num vertices & Number of vertices in each graph & $[5, 30]$ & 6 \\
edge prob & the edge probability of the Erdos-Renyi graph & $[0.1, 0.75]$ & 0.72\\
train prob & number of graphs in the dataset used for training & $[0.2, 0.6]$ & 0.60 \\
\bottomrule
\end{tabularx}
\caption{\label{tab:ss-values} Graph Property Prediction generator values used in \gw Mode 2 and Mode 3 experiments.}
\vspace{-4mm}
\end{table*}

%% file: sample-sigconf.bbl

\begin{thebibliography}{53}

%%% the Web address.

\ifx \showCODEN    \undefined \def \showCODEN     #1{\unskip}     \fi
\ifx \showDOI      \undefined \def \showDOI       #1{#1}\fi
\ifx \showISBNx    \undefined \def \showISBNx     #1{\unskip}     \fi
\ifx \showISBNxiii \undefined \def \showISBNxiii  #1{\unskip}     \fi
\ifx \showISSN     \undefined \def \showISSN      #1{\unskip}     \fi
\ifx \showLCCN     \undefined \def \showLCCN      #1{\unskip}     \fi
\ifx \shownote     \undefined \def \shownote      #1{#1}          \fi
\ifx \showarticletitle \undefined \def \showarticletitle #1{#1}   \fi
\ifx \showURL      \undefined \def \showURL       {\relax}        \fi
% invisible to TeX
\providecommand\bibfield[2]{#2}
\providecommand\bibinfo[2]{#2}
\providecommand\natexlab[1]{#1}
\providecommand\showeprint[2][]{arXiv:#2}

\bibitem[tfg(2021)]%
        {tfgnn}
 \bibinfo{year}{2021}\natexlab{}.
\newblock \bibinfo{title}{{Tensorflow GNN}}.
\newblock \bibinfo{howpublished}{\url{https://github.com/tensorflow/gnn}}.
\newblock


\bibitem[Abbe(2017)]%
        {abbe2017community}
\bibfield{author}{\bibinfo{person}{Emmanuel Abbe}.}
  \bibinfo{year}{2017}\natexlab{}.
\newblock \showarticletitle{Community detection and stochastic block models:
  recent developments}.
\newblock \bibinfo{journal}{\emph{JMLR}} (\bibinfo{year}{2017}).
\newblock


\bibitem[Adamic and Adar(2003)]%
        {adamic2003friends}
\bibfield{author}{\bibinfo{person}{Lada~A Adamic} {and} \bibinfo{person}{Eytan
  Adar}.} \bibinfo{year}{2003}\natexlab{}.
\newblock \showarticletitle{Friends and neighbors on the web}.
\newblock \bibinfo{journal}{\emph{Social networks}} (\bibinfo{year}{2003}).
\newblock


\bibitem[Bianchi et~al\mbox{.}(2021)]%
        {bianchi2021graph}
\bibfield{author}{\bibinfo{person}{Filippo~Maria Bianchi},
  \bibinfo{person}{Daniele Grattarola}, \bibinfo{person}{Lorenzo Livi}, {and}
  \bibinfo{person}{Cesare Alippi}.} \bibinfo{year}{2021}\natexlab{}.
\newblock \showarticletitle{Graph neural networks with convolutional arma
  filters}.
\newblock \bibinfo{journal}{\emph{IEEE Transactions on Pattern Analysis and
  Machine Intelligence}} (\bibinfo{year}{2021}).
\newblock


\bibitem[Brin and Page(1998)]%
        {brin1998anatomy}
\bibfield{author}{\bibinfo{person}{Sergey Brin} {and} \bibinfo{person}{Lawrence
  Page}.} \bibinfo{year}{1998}\natexlab{}.
\newblock \showarticletitle{The anatomy of a large-scale hypertextual web
  search engine}.
\newblock \bibinfo{journal}{\emph{Computer networks and ISDN systems}}
  (\bibinfo{year}{1998}).
\newblock


\bibitem[Brockschmidt(2020)]%
        {brockschmidt2020gnn}
\bibfield{author}{\bibinfo{person}{Marc Brockschmidt}.}
  \bibinfo{year}{2020}\natexlab{}.
\newblock \showarticletitle{Gnn-film: Graph neural networks with feature-wise
  linear modulation}. In \bibinfo{booktitle}{\emph{ICML}}. PMLR.
\newblock


\bibitem[Brody et~al\mbox{.}(2021)]%
        {brody2021attentive}
\bibfield{author}{\bibinfo{person}{Shaked Brody}, \bibinfo{person}{Uri Alon},
  {and} \bibinfo{person}{Eran Yahav}.} \bibinfo{year}{2021}\natexlab{}.
\newblock \showarticletitle{How Attentive are Graph Attention Networks?}
\newblock \bibinfo{journal}{\emph{arXiv preprint arXiv:2105.14491}}
  (\bibinfo{year}{2021}).
\newblock


\bibitem[Chami et~al\mbox{.}(2020)]%
        {chami2020machine}
\bibfield{author}{\bibinfo{person}{Ines Chami}, \bibinfo{person}{Sami
  Abu-El-Haija}, \bibinfo{person}{Bryan Perozzi}, \bibinfo{person}{Christopher
  R{\'e}}, {and} \bibinfo{person}{Kevin Murphy}.}
  \bibinfo{year}{2020}\natexlab{}.
\newblock \showarticletitle{Machine learning on graphs: A model and
  comprehensive taxonomy}.
\newblock \bibinfo{journal}{\emph{arXiv preprint arXiv:2005.03675}}
  (\bibinfo{year}{2020}).
\newblock


\bibitem[Chen et~al\mbox{.}(2020)]%
        {chen2020can}
\bibfield{author}{\bibinfo{person}{Zhengdao Chen}, \bibinfo{person}{Lei Chen},
  \bibinfo{person}{Soledad Villar}, {and} \bibinfo{person}{Joan Bruna}.}
  \bibinfo{year}{2020}\natexlab{}.
\newblock \showarticletitle{Can graph neural networks count substructures?}
\newblock \bibinfo{journal}{\emph{arXiv preprint arXiv:2002.04025}}
  (\bibinfo{year}{2020}).
\newblock


\bibitem[Cobbe et~al\mbox{.}(2020)]%
        {cobbe2020leveraging}
\bibfield{author}{\bibinfo{person}{Karl Cobbe}, \bibinfo{person}{Chris Hesse},
  \bibinfo{person}{Jacob Hilton}, {and} \bibinfo{person}{John Schulman}.}
  \bibinfo{year}{2020}\natexlab{}.
\newblock \showarticletitle{Leveraging procedural generation to benchmark
  reinforcement learning}. In \bibinfo{booktitle}{\emph{ICML}}.
\newblock


\bibitem[Dacrema et~al\mbox{.}(2019)]%
        {dacrema2019we}
\bibfield{author}{\bibinfo{person}{Maurizio~Ferrari Dacrema},
  \bibinfo{person}{Paolo Cremonesi}, {and} \bibinfo{person}{Dietmar Jannach}.}
  \bibinfo{year}{2019}\natexlab{}.
\newblock \showarticletitle{Are we really making much progress? A worrying
  analysis of recent neural recommendation approaches}. In
  \bibinfo{booktitle}{\emph{RecSys}}.
\newblock


\bibitem[Dice(1945)]%
        {dice1945measures}
\bibfield{author}{\bibinfo{person}{Lee~R Dice}.}
  \bibinfo{year}{1945}\natexlab{}.
\newblock \showarticletitle{Measures of the amount of ecologic association
  between species}.
\newblock \bibinfo{journal}{\emph{Ecology}} (\bibinfo{year}{1945}).
\newblock


\bibitem[Dwivedi et~al\mbox{.}(2020)]%
        {dwivedi2020benchmarking}
\bibfield{author}{\bibinfo{person}{Vijay~Prakash Dwivedi},
  \bibinfo{person}{Chaitanya~K Joshi}, \bibinfo{person}{Thomas Laurent},
  \bibinfo{person}{Yoshua Bengio}, {and} \bibinfo{person}{Xavier Bresson}.}
  \bibinfo{year}{2020}\natexlab{}.
\newblock \showarticletitle{Benchmarking graph neural networks}.
\newblock \bibinfo{journal}{\emph{arXiv preprint arXiv:2003.00982}}
  (\bibinfo{year}{2020}).
\newblock


\bibitem[Errica et~al\mbox{.}(2020)]%
        {errica2019fair}
\bibfield{author}{\bibinfo{person}{Federico Errica}, \bibinfo{person}{Marco
  Podda}, \bibinfo{person}{Davide Bacciu}, {and} \bibinfo{person}{Alessio
  Micheli}.} \bibinfo{year}{2020}\natexlab{}.
\newblock \showarticletitle{A fair comparison of graph neural networks for
  graph classification}. In \bibinfo{booktitle}{\emph{ICLR}}.
\newblock


\bibitem[Fey and Lenssen(2019)]%
        {Fey/Lenssen/2019}
\bibfield{author}{\bibinfo{person}{Matthias Fey} {and} \bibinfo{person}{Jan~E.
  Lenssen}.} \bibinfo{year}{2019}\natexlab{}.
\newblock \showarticletitle{Fast Graph Representation Learning with {PyTorch
  Geometric}}. In \bibinfo{booktitle}{\emph{ICLR Workshop on Representation
  Learning on Graphs and Manifolds}}.
\newblock


\bibitem[Grover and Leskovec(2016)]%
        {grover2016node2vec}
\bibfield{author}{\bibinfo{person}{Aditya Grover} {and} \bibinfo{person}{Jure
  Leskovec}.} \bibinfo{year}{2016}\natexlab{}.
\newblock \showarticletitle{node2vec: Scalable feature learning for networks}.
  In \bibinfo{booktitle}{\emph{KDD}}.
\newblock


\bibitem[Hamilton et~al\mbox{.}(2017)]%
        {hamilton2017inductive}
\bibfield{author}{\bibinfo{person}{William~L Hamilton}, \bibinfo{person}{Rex
  Ying}, {and} \bibinfo{person}{Jure Leskovec}.}
  \bibinfo{year}{2017}\natexlab{}.
\newblock \showarticletitle{Inductive representation learning on large graphs}.
  In \bibinfo{booktitle}{\emph{NIPS}}.
\newblock


\bibitem[Hu et~al\mbox{.}(2020)]%
        {hu2020open}
\bibfield{author}{\bibinfo{person}{Weihua Hu}, \bibinfo{person}{Matthias Fey},
  \bibinfo{person}{Marinka Zitnik}, \bibinfo{person}{Yuxiao Dong},
  \bibinfo{person}{Hongyu Ren}, \bibinfo{person}{Bowen Liu},
  \bibinfo{person}{Michele Catasta}, {and} \bibinfo{person}{Jure Leskovec}.}
  \bibinfo{year}{2020}\natexlab{}.
\newblock \showarticletitle{Open Graph Benchmark: Datasets for machine learning
  on graphs}. In \bibinfo{booktitle}{\emph{NeurIPS}}.
\newblock


\bibitem[Jaccard(1912)]%
        {jaccard1912distribution}
\bibfield{author}{\bibinfo{person}{Paul Jaccard}.}
  \bibinfo{year}{1912}\natexlab{}.
\newblock \showarticletitle{The distribution of the flora in the alpine zone}.
\newblock \bibinfo{journal}{\emph{New Phytologist}} (\bibinfo{year}{1912}).
\newblock


\bibitem[Karrer and Newman(2011)]%
        {karrer2011stochastic}
\bibfield{author}{\bibinfo{person}{Brian Karrer} {and} \bibinfo{person}{Mark~EJ
  Newman}.} \bibinfo{year}{2011}\natexlab{}.
\newblock \showarticletitle{Stochastic blockmodels and community structure in
  networks}.
\newblock \bibinfo{journal}{\emph{Physical review E}} \bibinfo{volume}{83},
  \bibinfo{number}{1} (\bibinfo{year}{2011}), \bibinfo{pages}{016107}.
\newblock


\bibitem[Khosla et~al\mbox{.}(2019)]%
        {khosla2019comparative}
\bibfield{author}{\bibinfo{person}{Megha Khosla}, \bibinfo{person}{Vinay
  Setty}, {and} \bibinfo{person}{Avishek Anand}.}
  \bibinfo{year}{2019}\natexlab{}.
\newblock \showarticletitle{A comparative study for unsupervised network
  representation learning}.
\newblock \bibinfo{journal}{\emph{TKDE}} (\bibinfo{year}{2019}).
\newblock


\bibitem[Kim and Oh(2021)]%
        {kim2021how}
\bibfield{author}{\bibinfo{person}{Dongkwan Kim} {and} \bibinfo{person}{Alice
  Oh}.} \bibinfo{year}{2021}\natexlab{}.
\newblock \showarticletitle{How to Find Your Friendly Neighborhood: Graph
  Attention Design with Self-Supervision}. In \bibinfo{booktitle}{\emph{ICLR}}.
\newblock


\bibitem[Kipf and Welling(2016)]%
        {kipf2016semi}
\bibfield{author}{\bibinfo{person}{Thomas~N Kipf} {and} \bibinfo{person}{Max
  Welling}.} \bibinfo{year}{2016}\natexlab{}.
\newblock \showarticletitle{Semi-supervised classification with graph
  convolutional networks}.
\newblock \bibinfo{journal}{\emph{arXiv preprint arXiv:1609.02907}}
  (\bibinfo{year}{2016}).
\newblock


\bibitem[Klicpera et~al\mbox{.}(2018)]%
        {klicpera2018predict}
\bibfield{author}{\bibinfo{person}{Johannes Klicpera},
  \bibinfo{person}{Aleksandar Bojchevski}, {and} \bibinfo{person}{Stephan
  G{\"u}nnemann}.} \bibinfo{year}{2018}\natexlab{}.
\newblock \showarticletitle{Predict then propagate: Graph neural networks meet
  personalized pagerank}.
\newblock \bibinfo{journal}{\emph{arXiv preprint arXiv:1810.05997}}
  (\bibinfo{year}{2018}).
\newblock


\bibitem[Koh et~al\mbox{.}(2021)]%
        {koh2021wilds}
\bibfield{author}{\bibinfo{person}{Pang~Wei Koh}, \bibinfo{person}{Shiori
  Sagawa}, \bibinfo{person}{Sang~Michael Xie}, \bibinfo{person}{Marvin Zhang},
  \bibinfo{person}{Akshay Balsubramani}, \bibinfo{person}{Weihua Hu},
  \bibinfo{person}{Michihiro Yasunaga}, \bibinfo{person}{Richard~Lanas
  Phillips}, \bibinfo{person}{Irena Gao}, \bibinfo{person}{Tony Lee},
  {et~al\mbox{.}}} \bibinfo{year}{2021}\natexlab{}.
\newblock \showarticletitle{Wilds: A benchmark of in-the-wild distribution
  shifts}. In \bibinfo{booktitle}{\emph{ICML}}.
\newblock


\bibitem[Leicht et~al\mbox{.}(2006)]%
        {leicht2006vertex}
\bibfield{author}{\bibinfo{person}{Elizabeth~A Leicht}, \bibinfo{person}{Petter
  Holme}, {and} \bibinfo{person}{Mark~EJ Newman}.}
  \bibinfo{year}{2006}\natexlab{}.
\newblock \showarticletitle{Vertex similarity in networks}.
\newblock \bibinfo{journal}{\emph{Physical Review E}} (\bibinfo{year}{2006}).
\newblock


\bibitem[Li et~al\mbox{.}(2021)]%
        {li2021deeperbiggerbetter}
\bibfield{author}{\bibinfo{person}{Guohao Li}, \bibinfo{person}{Jesus Zarzar},
  \bibinfo{person}{Hesham Mostafa}, \bibinfo{person}{Sohil Shah},
  \bibinfo{person}{Marcel Nassar}, \bibinfo{person}{Daniel Cummings},
  \bibinfo{person}{Sami Abu-El-Haija}, \bibinfo{person}{Bernard Ghanem}, {and}
  \bibinfo{person}{Matthias M{\"u}ller}.} \bibinfo{year}{2021}\natexlab{}.
\newblock \showarticletitle{DEEPERBIGGERBETTER for OGB-LSC at KDD cup 2021}.
\newblock  (\bibinfo{year}{2021}).
\newblock


\bibitem[Lipton and Steinhardt(2018)]%
        {lipton2018troubling}
\bibfield{author}{\bibinfo{person}{Zachary~C Lipton} {and}
  \bibinfo{person}{Jacob Steinhardt}.} \bibinfo{year}{2018}\natexlab{}.
\newblock \showarticletitle{Troubling trends in machine learning scholarship}.
\newblock \bibinfo{journal}{\emph{arXiv preprint arXiv:1807.03341}}
  (\bibinfo{year}{2018}).
\newblock


\bibitem[Luzhnica et~al\mbox{.}(2019)]%
        {luzhnica2019graph}
\bibfield{author}{\bibinfo{person}{Enxhell Luzhnica}, \bibinfo{person}{Ben
  Day}, {and} \bibinfo{person}{Pietro Li{\`o}}.}
  \bibinfo{year}{2019}\natexlab{}.
\newblock \showarticletitle{On graph classification networks, datasets and
  baselines}.
\newblock \bibinfo{journal}{\emph{arXiv preprint arXiv:1905.04682}}
  (\bibinfo{year}{2019}).
\newblock


\bibitem[Mara et~al\mbox{.}(2021)]%
        {marareproducible}
\bibfield{author}{\bibinfo{person}{Alexandru Mara}, \bibinfo{person}{Jefrey
  Lijffijt}, {and} \bibinfo{person}{Tijl de Bie}.}
  \bibinfo{year}{2021}\natexlab{}.
\newblock \showarticletitle{Reproducible Evaluations of Network Representation
  Learning Models Using EvalNE}.
\newblock \bibinfo{journal}{\emph{WWW'21, Workshop on Graph Learning
  Benchmarks}} (\bibinfo{year}{2021}).
\newblock


\bibitem[Morris et~al\mbox{.}(2020)]%
        {morris2020tudataset}
\bibfield{author}{\bibinfo{person}{Christopher Morris}, \bibinfo{person}{Nils~M
  Kriege}, \bibinfo{person}{Franka Bause}, \bibinfo{person}{Kristian Kersting},
  \bibinfo{person}{Petra Mutzel}, {and} \bibinfo{person}{Marion Neumann}.}
  \bibinfo{year}{2020}\natexlab{}.
\newblock \showarticletitle{Tudataset: A collection of benchmark datasets for
  learning with graphs}.
\newblock \bibinfo{journal}{\emph{arXiv preprint arXiv:2007.08663}}
  (\bibinfo{year}{2020}).
\newblock


\bibitem[Nie et~al\mbox{.}(2019)]%
        {nie2019adversarial}
\bibfield{author}{\bibinfo{person}{Yixin Nie}, \bibinfo{person}{Adina
  Williams}, \bibinfo{person}{Emily Dinan}, \bibinfo{person}{Mohit Bansal},
  \bibinfo{person}{Jason Weston}, {and} \bibinfo{person}{Douwe Kiela}.}
  \bibinfo{year}{2019}\natexlab{}.
\newblock \showarticletitle{Adversarial NLI: A new benchmark for natural
  language understanding}.
\newblock \bibinfo{journal}{\emph{arXiv preprint arXiv:1910.14599}}
  (\bibinfo{year}{2019}).
\newblock


\bibitem[Ravasz et~al\mbox{.}(2002)]%
        {ravasz2002hierarchical}
\bibfield{author}{\bibinfo{person}{Erzs{\'e}bet Ravasz},
  \bibinfo{person}{Anna~Lisa Somera}, \bibinfo{person}{Dale~A Mongru},
  \bibinfo{person}{Zolt{\'a}n~N Oltvai}, {and} \bibinfo{person}{A-L
  Barab{\'a}si}.} \bibinfo{year}{2002}\natexlab{}.
\newblock \showarticletitle{Hierarchical organization of modularity in
  metabolic networks}.
\newblock \bibinfo{journal}{\emph{Science}} (\bibinfo{year}{2002}).
\newblock


\bibitem[Recht et~al\mbox{.}(2018)]%
        {recht2018cifar}
\bibfield{author}{\bibinfo{person}{Benjamin Recht}, \bibinfo{person}{Rebecca
  Roelofs}, \bibinfo{person}{Ludwig Schmidt}, {and} \bibinfo{person}{Vaishaal
  Shankar}.} \bibinfo{year}{2018}\natexlab{}.
\newblock \showarticletitle{Do cifar-10 classifiers generalize to cifar-10?}
\newblock \bibinfo{journal}{\emph{arXiv preprint arXiv:1806.00451}}
  (\bibinfo{year}{2018}).
\newblock


\bibitem[Rossi and Ahmed(2015)]%
        {networkrepository}
\bibfield{author}{\bibinfo{person}{Ryan~A. Rossi} {and}
  \bibinfo{person}{Nesreen~K. Ahmed}.} \bibinfo{year}{2015}\natexlab{}.
\newblock \showarticletitle{The Network Data Repository with Interactive Graph
  Analytics and Visualization}. In \bibinfo{booktitle}{\emph{AAAI}}.
\newblock
\urldef\tempurl%
\url{https://networkrepository.com}
\showURL{%
\tempurl}


\bibitem[Shah(2020)]%
        {shah2020scale}
\bibfield{author}{\bibinfo{person}{Neil Shah}.}
  \bibinfo{year}{2020}\natexlab{}.
\newblock \showarticletitle{Scale-Free, Attributed and Class-Assortative Graph
  Generation to Facilitate Introspection of Graph Neural Networks}.
\newblock \bibinfo{journal}{\emph{WWW'21, Workshop on Graph Learning
  Benchmarks}} (\bibinfo{year}{2020}).
\newblock


\bibitem[Shchur et~al\mbox{.}(2018)]%
        {shchur2018pitfalls}
\bibfield{author}{\bibinfo{person}{Oleksandr Shchur},
  \bibinfo{person}{Maximilian Mumme}, \bibinfo{person}{Aleksandar Bojchevski},
  {and} \bibinfo{person}{Stephan G{\"u}nnemann}.}
  \bibinfo{year}{2018}\natexlab{}.
\newblock \showarticletitle{Pitfalls of graph neural network evaluation}.
\newblock \bibinfo{journal}{\emph{arXiv preprint arXiv:1811.05868}}
  (\bibinfo{year}{2018}).
\newblock


\bibitem[Shi et~al\mbox{.}(2021)]%
        {shi2020masked}
\bibfield{author}{\bibinfo{person}{Yunsheng Shi}, \bibinfo{person}{Zhengjie
  Huang}, \bibinfo{person}{Shikun Feng}, \bibinfo{person}{Hui Zhong},
  \bibinfo{person}{Wenjin Wang}, {and} \bibinfo{person}{Yu Sun}.}
  \bibinfo{year}{2021}\natexlab{}.
\newblock \showarticletitle{Masked label prediction: Unified message passing
  model for semi-supervised classification}.
\newblock  (\bibinfo{year}{2021}).
\newblock


\bibitem[Shridhar et~al\mbox{.}(2020)]%
        {shridhar2020alfred}
\bibfield{author}{\bibinfo{person}{Mohit Shridhar}, \bibinfo{person}{Jesse
  Thomason}, \bibinfo{person}{Daniel Gordon}, \bibinfo{person}{Yonatan Bisk},
  \bibinfo{person}{Winson Han}, \bibinfo{person}{Roozbeh Mottaghi},
  \bibinfo{person}{Luke Zettlemoyer}, {and} \bibinfo{person}{Dieter Fox}.}
  \bibinfo{year}{2020}\natexlab{}.
\newblock \showarticletitle{Alfred: A benchmark for interpreting grounded
  instructions for everyday tasks}. In \bibinfo{booktitle}{\emph{CVPR}}.
\newblock


\bibitem[Snedecor(1957)]%
        {snedecor1957statistical}
\bibfield{author}{\bibinfo{person}{George~W Snedecor}.}
  \bibinfo{year}{1957}\natexlab{}.
\newblock \bibinfo{title}{Statistical methods}.
\newblock
\newblock


\bibitem[S{\o}rensen(1948)]%
        {sorensen1948method}
\bibfield{author}{\bibinfo{person}{Th~A S{\o}rensen}.}
  \bibinfo{year}{1948}\natexlab{}.
\newblock \showarticletitle{A method of establishing groups of equal amplitude
  in plant sociology based on similarity of species content and its application
  to analyses of the vegetation on Danish commons}.
\newblock \bibinfo{journal}{\emph{Biol. Skar.}} (\bibinfo{year}{1948}).
\newblock


\bibitem[Thorndike(1953)]%
        {thorndike1953belongs}
\bibfield{author}{\bibinfo{person}{Robert~L Thorndike}.}
  \bibinfo{year}{1953}\natexlab{}.
\newblock \showarticletitle{Who belongs in the family?}
\newblock \bibinfo{journal}{\emph{Psychometrika}} (\bibinfo{year}{1953}).
\newblock


\bibitem[Tsitsulin et~al\mbox{.}(2020)]%
        {tsitsulin2020graph}
\bibfield{author}{\bibinfo{person}{Anton Tsitsulin}, \bibinfo{person}{John
  Palowitch}, \bibinfo{person}{Bryan Perozzi}, {and} \bibinfo{person}{Emmanuel
  M{\"u}ller}.} \bibinfo{year}{2020}\natexlab{}.
\newblock \showarticletitle{Graph clustering with graph neural networks}.
\newblock \bibinfo{journal}{\emph{arXiv preprint arXiv:2006.16904}}
  (\bibinfo{year}{2020}).
\newblock


\bibitem[Tsitsulin et~al\mbox{.}(2021)]%
        {tsitsulin2021synthetic}
\bibfield{author}{\bibinfo{person}{Anton Tsitsulin}, \bibinfo{person}{Benedek
  Rozemberczki}, \bibinfo{person}{John Palowitch}, {and} \bibinfo{person}{Bryan
  Perozzi}.} \bibinfo{year}{2021}\natexlab{}.
\newblock \showarticletitle{Synthetic Graph Generation to Benchmark Graph
  Learning}.
\newblock \bibinfo{journal}{\emph{WWW'21, Workshop on Graph Learning
  Benchmarks}} (\bibinfo{year}{2021}).
\newblock


\bibitem[Veli{\v{c}}kovi{\'c} et~al\mbox{.}(2017)]%
        {velivckovic2017graph}
\bibfield{author}{\bibinfo{person}{Petar Veli{\v{c}}kovi{\'c}},
  \bibinfo{person}{Guillem Cucurull}, \bibinfo{person}{Arantxa Casanova},
  \bibinfo{person}{Adriana Romero}, \bibinfo{person}{Pietro Lio}, {and}
  \bibinfo{person}{Yoshua Bengio}.} \bibinfo{year}{2017}\natexlab{}.
\newblock \showarticletitle{Graph attention networks}.
\newblock \bibinfo{journal}{\emph{arXiv preprint arXiv:1710.10903}}
  (\bibinfo{year}{2017}).
\newblock


\bibitem[Wolfson et~al\mbox{.}(2020)]%
        {wolfson2020break}
\bibfield{author}{\bibinfo{person}{Tomer Wolfson}, \bibinfo{person}{Mor Geva},
  \bibinfo{person}{Ankit Gupta}, \bibinfo{person}{Matt Gardner},
  \bibinfo{person}{Yoav Goldberg}, \bibinfo{person}{Daniel Deutch}, {and}
  \bibinfo{person}{Jonathan Berant}.} \bibinfo{year}{2020}\natexlab{}.
\newblock \showarticletitle{Break it down: A question understanding benchmark}.
\newblock \bibinfo{journal}{\emph{ACL}} (\bibinfo{year}{2020}).
\newblock


\bibitem[Wu et~al\mbox{.}(2019)]%
        {wu2019simplifying}
\bibfield{author}{\bibinfo{person}{Felix Wu}, \bibinfo{person}{Amauri Souza},
  \bibinfo{person}{Tianyi Zhang}, \bibinfo{person}{Christopher Fifty},
  \bibinfo{person}{Tao Yu}, {and} \bibinfo{person}{Kilian Weinberger}.}
  \bibinfo{year}{2019}\natexlab{}.
\newblock \showarticletitle{Simplifying graph convolutional networks}. In
  \bibinfo{booktitle}{\emph{ICML}}. PMLR.
\newblock


\bibitem[Wu et~al\mbox{.}(2020)]%
        {wu2020comprehensive}
\bibfield{author}{\bibinfo{person}{Zonghan Wu}, \bibinfo{person}{Shirui Pan},
  \bibinfo{person}{Fengwen Chen}, \bibinfo{person}{Guodong Long},
  \bibinfo{person}{Chengqi Zhang}, {and} \bibinfo{person}{S~Yu Philip}.}
  \bibinfo{year}{2020}\natexlab{}.
\newblock \showarticletitle{A comprehensive survey on graph neural networks}.
\newblock \bibinfo{journal}{\emph{IEEE transactions on neural networks and
  learning systems}} (\bibinfo{year}{2020}).
\newblock


\bibitem[Xu et~al\mbox{.}(2018)]%
        {xu2018powerful}
\bibfield{author}{\bibinfo{person}{Keyulu Xu}, \bibinfo{person}{Weihua Hu},
  \bibinfo{person}{Jure Leskovec}, {and} \bibinfo{person}{Stefanie Jegelka}.}
  \bibinfo{year}{2018}\natexlab{}.
\newblock \showarticletitle{How powerful are graph neural networks?}
\newblock \bibinfo{journal}{\emph{arXiv preprint arXiv:1810.00826}}
  (\bibinfo{year}{2018}).
\newblock


\bibitem[You et~al\mbox{.}(2020)]%
        {you2020design}
\bibfield{author}{\bibinfo{person}{Jiaxuan You}, \bibinfo{person}{Zhitao Ying},
  {and} \bibinfo{person}{Jure Leskovec}.} \bibinfo{year}{2020}\natexlab{}.
\newblock \showarticletitle{Design space for graph neural networks}. In
  \bibinfo{booktitle}{\emph{NeurIPS}}.
\newblock


\bibitem[Zhao et~al\mbox{.}(2020)]%
        {zhao2020pipeline}
\bibfield{author}{\bibinfo{person}{Wentao Zhao}, \bibinfo{person}{Dalin Zhou},
  \bibinfo{person}{Xinguo Qiu}, {and} \bibinfo{person}{Wei Jiang}.}
  \bibinfo{year}{2020}\natexlab{}.
\newblock \showarticletitle{A pipeline for fair comparison of graph neural
  networks in node classification tasks}.
\newblock \bibinfo{journal}{\emph{arXiv preprint arXiv:2012.10619}}
  (\bibinfo{year}{2020}).
\newblock


\bibitem[Zhou et~al\mbox{.}(2009)]%
        {zhou2009predicting}
\bibfield{author}{\bibinfo{person}{Tao Zhou}, \bibinfo{person}{Linyuan L{\"u}},
  {and} \bibinfo{person}{Yi-Cheng Zhang}.} \bibinfo{year}{2009}\natexlab{}.
\newblock \showarticletitle{Predicting missing links via local information}.
\newblock \bibinfo{journal}{\emph{The European Physical Journal B}}
  (\bibinfo{year}{2009}).
\newblock


\bibitem[Zhu et~al\mbox{.}(2020)]%
        {zhu2020beyond}
\bibfield{author}{\bibinfo{person}{Jiong Zhu}, \bibinfo{person}{Yujun Yan},
  \bibinfo{person}{Lingxiao Zhao}, \bibinfo{person}{Mark Heimann},
  \bibinfo{person}{Leman Akoglu}, {and} \bibinfo{person}{Danai Koutra}.}
  \bibinfo{year}{2020}\natexlab{}.
\newblock \showarticletitle{Beyond homophily in graph neural networks: Current
  limitations and effective designs}.
\newblock \bibinfo{journal}{\emph{arXiv preprint arXiv:2006.11468}}
  (\bibinfo{year}{2020}).
\newblock


\end{thebibliography}
